\documentclass[lettersize,journal]{IEEEtran}
\usepackage{amsmath,amsfonts}
\usepackage{algorithmic}
\usepackage{algorithm}
\usepackage{array}
\usepackage[caption=false,font=normalsize,labelfont=sf,textfont=sf]{subfig}
\usepackage{textcomp}
\usepackage{stfloats}
\usepackage{url}
\usepackage{verbatim}
\usepackage{graphicx}
\usepackage{cite}
\usepackage[table,xcdraw]{xcolor}
\usepackage{bbm}
\usepackage{multirow}
\usepackage{booktabs}
\usepackage{bbding}
\usepackage{array}
\begin{document}

\title{Evolved Hierarchical Masking for  Self-Supervised \\ Learning}

\author{Zhanzhou Feng, Shiliang Zhang~\IEEEmembership{Senior Member, IEEE}
  \thanks{Zhanzhou Feng and Shiliang Zhang are with National Key Laboratory for Multimedia Information Processing, School of Computer Science, Peking University, Beijing 100871, China (e-mail: fengzz@stu.pku.edu.cn; slzhang.jdl@pku.edu.cn).
  }
}

\markboth{Journal of \LaTeX\ Class Files,~Vol.~14, No.~8, August~2021}%
{Shell \MakeLowercase{\textit{et al.}}: A Sample Article Using IEEEtran.cls for IEEE Journals}


\maketitle

\begin{abstract}
  Existing Masked Image Modeling methods apply fixed mask patterns to guide the self-supervised training.
  As those mask patterns resort to different criteria to depict image contents, sticking to a fixed pattern leads to a limited vision cues modeling capability.
  This paper introduces an evolved hierarchical masking method to pursue general visual cues modeling in self-supervised learning.
  The proposed method leverages the vision model being trained to parse the input visual cues into a hierarchy structure, which is hence adopted to generate masks accordingly.
  The accuracy of hierarchy is on par with the capability of the model being trained, leading to evolved mask patterns at different training stages.
  Initially, generated masks focus on low-level visual cues to grasp basic textures, then gradually evolve to depict higher-level cues to reinforce the learning of more complicated object semantics and contexts. Our method does not require extra pre-trained models or annotations and ensures training efficiency by evolving the training difficulty.
  We conduct extensive experiments on seven downstream tasks including partial-duplicate image retrieval relying on low-level details, as well as image classification and semantic segmentation that require semantic parsing capability. Experimental results demonstrate that it substantially boosts performance across these tasks.
  For instance, it surpasses the recent MAE by 1.1\% in imageNet-1K classification and 1.4\% in ADE20K segmentation with the same training epochs. We also align the proposed method with the current research focus on LLMs. The proposed approach bridges the gap with large-scale pre-training on semantic demanding tasks and enhances intricate detail perception in tasks requiring low-level feature recognition.

\end{abstract}


\section{Introduction}
\IEEEPARstart{R}{ecent} years have witnessed a boom in continuously growing representation learning capability and data demands of deep neural networks.
To tackle the increasing demand for labeled data, Self-Supervised Learning (SSL) has emerged as a promising paradigm in pre-training deep models.
By constructing pretext training tasks, the pretrained model can acquire transferable representations from unlabeled large-scale data.
In the field of natural language processing, Masked Language Modeling (MLM) has emerged as a predominant pre-training task, constituting an essential component in the training process of renowned models such as BERT~\cite{devlin2018bert} and GPT~\cite{brown2020language}.
MLM masks several words in the input sentences and supervises the network to recover masked words according to the semantics provided by the remaining words.

Inspired by the success of MLM, a vision counterpart, Masked Image Modeling (MIM) is proposed to pre-train vision models on unlabeled images.
MIM follows a similar idea from language to mask a portion of input image patches, then the pretext task is to recover masked contents from visible patches.
As images are not structured like sentences, different MIM works have to resort to different criteria to generate mask patterns. We categorize mask patterns in existing works into three based on their followed masking criteria.
Some works like MAE~\cite{he2022masked} and SimMIM~\cite{xie2022simmim}
randomly mask image patches, assuming input information density is uniformly distributed.
Another line of works, like MST~\cite{li2021mst}, aims to preserve crucial cues in the image to enhance the learning of local context.
A third line, including AttnMask~\cite{kakogeorgiou2022hide}, I-JEPA~\cite{assran2023self}, and SemMAE~\cite{li2022semmae}, suggests completely masking cues such as object regions in images to pose a more challenging learning task.

\begin{figure}[t]
  \centering
   \includegraphics[width=0.92\linewidth]{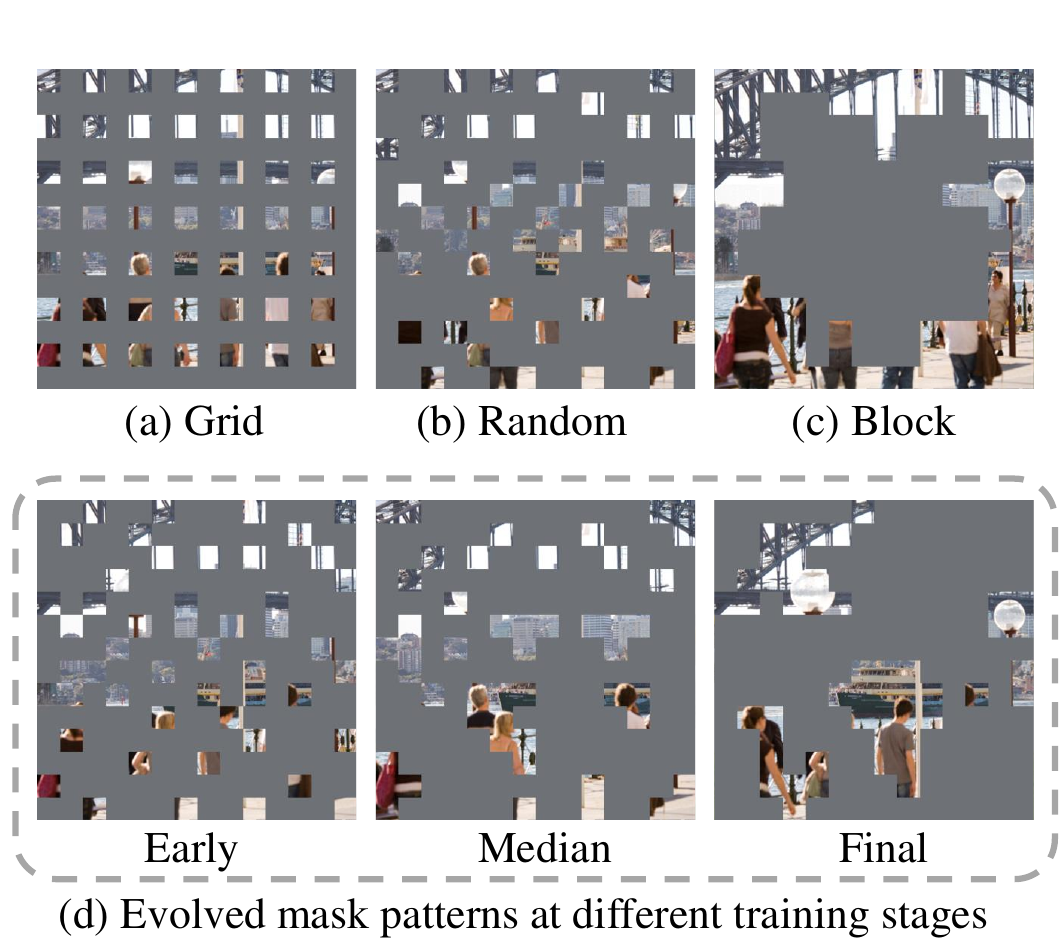}

   \caption{(a), (b), and (c) are three basic mask patterns adopted in existing MIM methods.
    (d) illustrates the proposed evolved hierarchical masking, where the generated mask patterns evolve with the capability of the vision model being trained.
   }
   \label{fig:teaser}
\end{figure}

\begin{figure}
  \centering
   \includegraphics[width=1.0\linewidth]{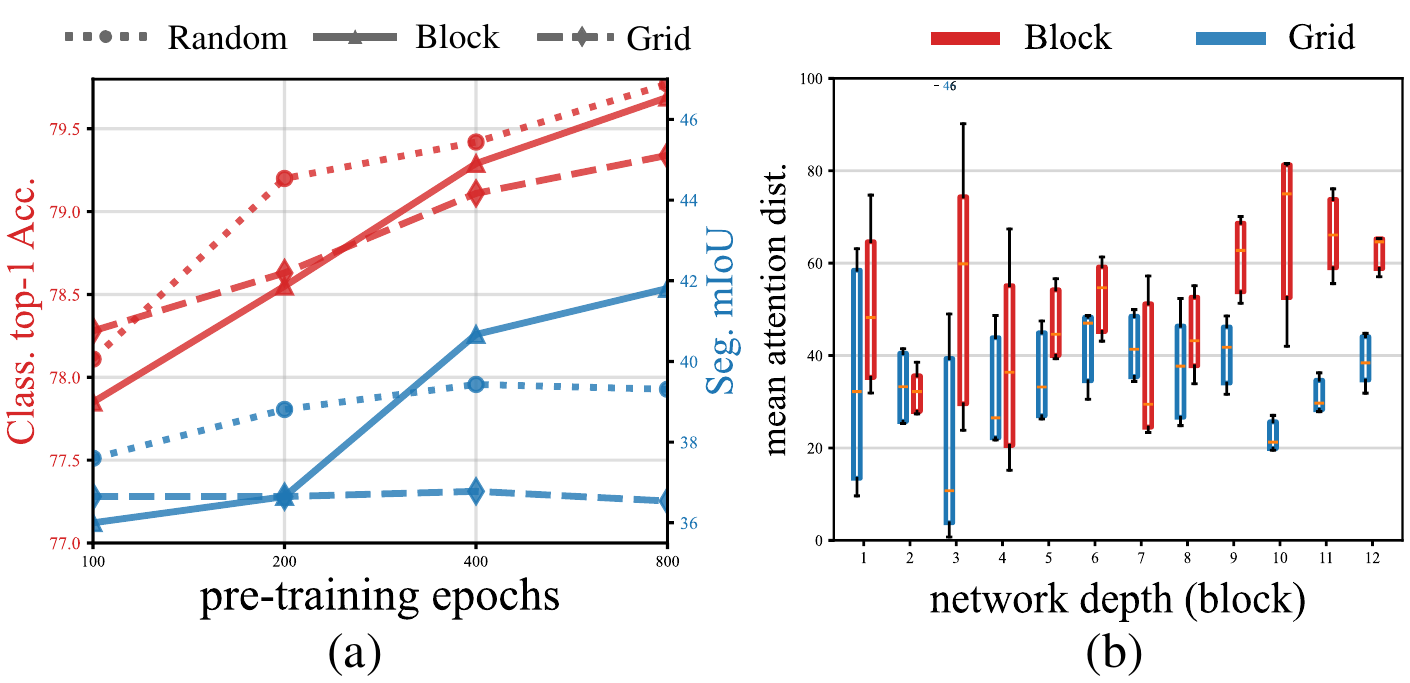}
   \caption{Illustration of effects of different mask patterns to downstream tasks in (a), and learned parameters in (b), respectively. In (a), random pattern and block pattern perform best in image classification and semantic segmentation, respectively. (b) shows the mean attention distance across images at different layers of the pre-trained model. The unit of the y-axis is the pixel. Those results indicate different mask patterns are suited to different tasks.
   }
   \label{fig:intro_curve}
\end{figure}

To study the impact of mask patterns on self-supervised pre-training, we test the performance of three commonly used masking methods, i.e., random masks, grid masks and block masks, as shown in Fig.~\ref{fig:teaser}(a)-(c). Fig.~\ref{fig:intro_curve}(a) explores their effects on two vision tasks. It can be observed that, random pattern and block pattern perform best in image classification and semantic segmentation, respectively. It is also clear that, more training epochs do not boost the performance of grid pattern and random pattern in segmentation. Fig.~\ref{fig:intro_curve}(b) further visualizes the average attention length of neurons at each layer of the pre-trained model.
We compute attention distance by averaging the distance between the query tokens and all other tokens, weighted by the attention weight following~\cite{dosovitskiy2020image}.
It indicates that, neurons trained by grid mask mostly focus on nearby regions with shorter attention distances. As longer attention distance benefits the learning of contextual cues, block pattern is more preferred by dense prediction tasks like semantic segmentation.

The above experiments indicate that, the choice of patterns for generating masks largely determines visual dependencies that the network could learn in pre-training.
For instance, masking the entire object regions enforces the network to learn correlated semantics and contexts of visible cues, which are helpful for image classification and object segmentation. On the other hand, grid mask patterns supervises the network neuron to attend to nearby regions, which are important for recovering low-level details and enhancing efficiency in learning tasks. Therefore, different mask patterns are suitable for different downstream tasks. This poses a challenge for self-supervised learning, i.e., the pre-training procedure has no clue which task it will be applied to. In other words, visual knowledge learned with fixed masking patterns is limited and hard to generalize to various downstream tasks.


\begin{figure}[t]
  \centering
   \includegraphics[width=1.0\linewidth]{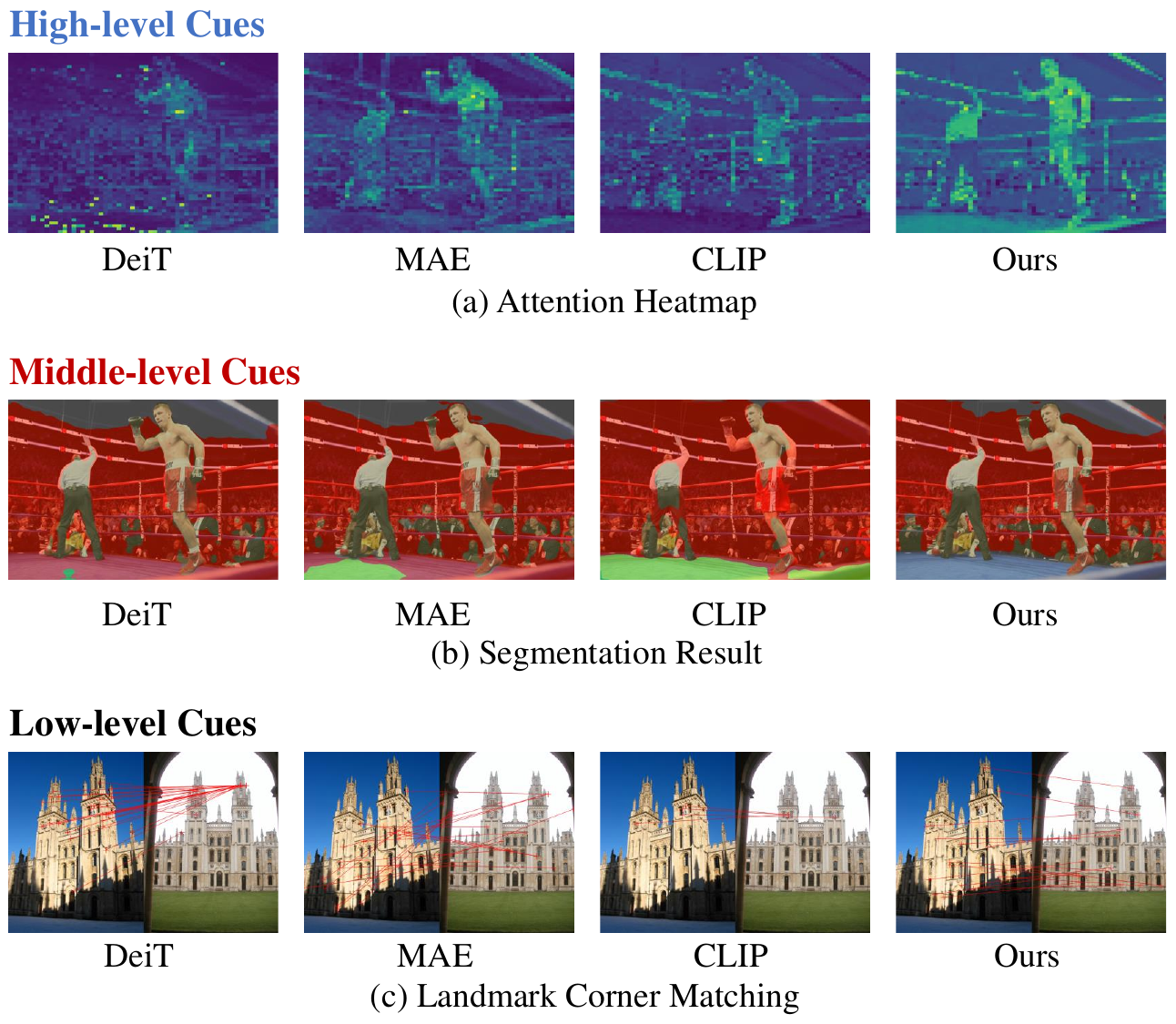}

   \caption{ Visualization of [CLS] attention heatmap from classification~\cite{deng2009imagenet} in (a), semantic segmentation result~\cite{zhou2017scene} in (b),  and landmark retrieval results on Oxford Building~\cite{philbin2007object,philbin2008lost} in (c). These visualizations show the proposed method exhibits superior visual cue acquisition capacity at different semantic levels. Best viewing with zoom-in.
   }
   \label{fig:teaser_visual}
\end{figure}

Mask pattern adopted by MLM is hence expected to enforce the learning with various visual cues. We propose to parse image contents into a hierarchy representing different levels of visual cues, and progressively mask visual cues according to this hierarchy. An adaptive part partition module is adopted to leverage the vision model being trained to construct the hierarchy, and generate mask patterns.
Establishment of the hierarchy starts by treating each image patch as a leaf node in a tree. We take the attention map produced by the current model as the similarity matrix among image patches. A pair of most similar nodes are hence merged to generate their parent node. 
This procedure continues until a single root encompasses all leaf nodes, which leads to a structured hierarchy representing visual cues at various scales.
It`s worth noting that the hierarchy is dynamically constructed each time an image is input into the system; thus, it evolves alongside the model training. 
As the model progressively learns new visual information, the generated hierarchical structure also improves.

We then perform masking by removing image patches on a certain level of the established hierarchy. Initially, masked patches are sampled from low-level hierarchies to depict visual details. This enforces the network to learn low-level visual cues, which are prerequisites for developing higher-level representations. As the training progresses, the masked location gradually advances to higher hierarchies to depict more complicated regions as illustrated in Fig.~\ref{fig:teaser}(d). In other words, this transition from low to high hierarchy generates simple patterns at the initial training stage, which hence evolves to eliminate accurate object parts to reinforce the learning of object semantics and contexts. This procedure progressively masks visual cues and reveals the underlying visual dependencies, e.g., from local textures to object parts and general semantics, to gradationally promote the representation learning capabilities. 


We test the proposed method on various downstream tasks requiring different visual cues modeling capabilities. As illustrated in Fig.~\ref{fig:teaser_visual} (a), the attention map of our method focuses on both the foreground player and the background referee, indicating a more comprehensive understanding of the boxing match scene. Fig.~\ref{fig:teaser_visual}(b) shows that our method performs better than previous works in image segmentation, showing its boosted performance in inferring region and pixel-level semantics. Fig.~\ref{fig:teaser_visual}(c) presents the results of image matching, which heavily relies on the discriminative power of local features extracted at corner points. It is clear that, our method gets the most reliable feature matching result among competitors.

Extensive experimental results also demonstrate that the proposed method produces a stronger initialization model for various downstream tasks.
We test the effectiveness of the proposed method on three popular MIM architectures, \emph{i.e.,} MAE~\cite{he2022masked}, BEiT~\cite{bao2021beit} and SimMIM~\cite{xie2022simmim}. Our method brings significant performance enhancement for those three architectures, especially on the semantic segmentation task, \emph{e.g.}, boosts the mIoU by $+1.42\%$, $+1.39\%$ and $+1.32\%$, respectively.
When compared with recent self-supervised learning methods, our method achieves comparable performance with fewer pre-training epochs, and superior performance with similar training epochs.
When combined the proposed method with Large Language Models, it bridges the performance gap between self-supervised techniques and large-scale pre-training in semantic demanding VQAv2~\cite{goyal2017making} task by $+2.2\%$.
Additionally, it amplifies the strength of self-supervised methods in recognizing intricate details.

To the best of our knowledge, this is an original effort on evolved hierarchy masking for self-supervised learning. It builds a semantic hierarchy to generate masks, leading to an evolved masking criteria at different training stages. Our method does not require extra pre-trained models or annotations. It effectively ensures training efficiency, and enhances the generalization ability of the trained model. As shown in various downstream tasks, our method shows better potentials to boost the performance of pre-trained vision models.

\section{Related works}
This work is closely related to self-supervised learning and masked image modeling. This section briefly reviews related works in recent years.

\textbf{Self-supervised Learning.}
Generally,  SSL  involves creating an annotation-free pretext task that helps pre-trained models learn visual representations by understanding connections in visual inputs.
Existing SSL methods can be divided into two categories based on how annotation-free tasks are designed.

The first category, contrastive learning, has dominated the learning algorithms in those works for the past few years.
Under an instance discrimination setting, each image is considered as an individual class~\cite{hadsell2006dimensionality}.
It works by pulling positive samples together and pushing negative samples apart.
A common technique involves optimizing an encoder to generate comparable embeddings for multiple perspectives of the same image. These perspectives are usually created using a range of carefully designed data augmentations, including random  cropping~\cite{chen2020simple}, horizontal flipping~\cite{wang2021dense}, and color adjustments~\cite{chen2020improved}, among others~\cite{zhang2022rethinking}.
However, this technique requires prior knowledge of task-specific invariances, which is often unknown during the pretraining phase.
Related works includes SimCLR~\cite{chen2020simple}, MoCo~\cite{he2020momentum}, Swav~\cite{caron2020unsupervised}, and BYOL~\cite{grill2020bootstrap}, \emph{etc.}
DINO~\cite{caron2021emerging} formulates self-supervised contrastive as self-distillation. The student model is trained to predict the output of a teacher network built with a momentum encoder.

Another category of SSL research predicts the original image based on the partially observed data.
For example, RotNet~\cite{gidaris2018unsupervised} predicts the 2D rotation applied to the input image.
The CFN~\cite{noroozi2016unsupervised} randomly shuffles the image patches and takes Jigsaw puzzles as the pretext task.
Autoencoder is a commonly used generative SSL model, which is trained by minimizing the reconstruction error.
It has an encoder that maps the input data to a latent space and a decoder to reconstruct the image conditioned on the latent representation.
Denoising autoencoders (DAE)~\cite{vincent2008extracting} corrupts input signal with random noise and context encoders regress an entire image region based on its surroundings.
A series of methods can be viewed as generalized DAE with different ways of generating corrupted images, including degrading the resolution~\cite{chen2020generative}, masking regions~\cite{pathak2016context}, or removing certain color channels~\cite{zhang2016colorful}, etc.
MIM also can be regarded as one of DAE variants.

\textbf{Masked Image Modeling.}
Inspired by the success of MLM~\cite{devlin2018bert,brown2020language} in NLP, MIM has been proposed to tackle the data-hungry issue of vision transformers~\cite{touvron2021training,dosovitskiy2020image}.
As one of the core designs of MIM, the mask methods largely determine the knowledge that the network acquires in the pre-training phase.
According to the masking criteria, this work divides existing MIM methods into three categories and discusses their characteristics and trade-offs.

The first category involves random masking, a widely adopted and straightforward approach.
MAE~\cite{he2022masked} is one of the representative works that utilize an asymmetric autoencoder to recover a randomly masked input.
SimMIM~\cite{xie2022simmim} randomly masks larger square patches and minimizes the $\ell_1$ loss between raw pixel values and predicted results.
The advantage of these methods lies in their simplicity and ease of implementation.
However, due to the complete randomness of the generated masks, extensive training iterations are required to ensure a comprehensive coverage of visual dependencies conveyed from the generated masks.
This inefficiency further increases the burden of computing resources during pre-training.

The second category prioritizes reserving crucial cues within the masking process.
For example, MST~\cite{li2021mst} masks only nonessential patches and preserves key patterns in images.
MFM~\cite{xie2022masked} uses low-pass/high-pass filters to perform masking, and most object regions with clear semantics are preserved.
These works preserve the most important contents in the image, posing an easier MIM task.
They facilitate rapid and stable convergence, enhance training efficiency, and effectively capture low-level visual details.
The limitation of this masking pattern lies in excessive shared information between masked and visible patches.
This can lead the network to mimic visible patches for content reconstruction, rather than developing high-level semantic understanding, resulting in subpar high-level semantic learning.


The third category proposes to mask clues like object regions completely. BEiT~\cite{bao2021beit} employs a block-wise masking method to mask some image objects as a whole. AttnMask~\cite{kakogeorgiou2022hide} proposes to mask patches belonging to the most attended objects. SemMAE~\cite{li2022semmae} leverages the iBOT~\cite{zhou2021ibot} for semantic segmentation and produces the mask according to the segmentation result. Besides, ADIOS~\cite{shi2022adversarial} utilizes a learned adversarial masking subnet to pose a more challenging MIM task.
Mahmoud's proposed masking strategy in I-JEPA~\cite{assran2023self} emphasizes the importance of selecting target blocks of sufficiently large scale and using a context block that provides ample information.
This category excels in modeling both high-level semantics, yielding top performance in several downstream tasks.
However, as most pre-trained networks start with randomly initialized parameters, there is a deficiency in modeling low-level visual cues.
In the absence of essential low-level details such as textures and edges, the deduced high-level semantics are more difficult to learn, resulting in convergence challenges and even potential training collapses.
Compared to SemMAE~\cite{li2022semmae} and ADIOS~\cite{shi2022adversarial}, our method does not introduce extra networks or training costs. It leverages the model being trained to determine the cues that should be masked. Evolved masks make training difficulty on par with the capability of the network being trained, hence ensuring a more effective and fast self-supervised learning.

\textbf{Multimodal LLMs.}
Large Language Models (LLMs)~\cite{zheng2023judging,liu2023visual} have demonstrated remarkable performance in various tasks without task-specific fine-tuning. Recently, researchers have explored combining LLMs with visual encoders to tackle multi-modal and even pure visual tasks, such as MiniGPT-4~\cite{zhu2023minigpt} and BLIP-2~\cite{li2023blip}. Under this framework, a visual tokenizer encodes the image, followed by bridges like MLP or Perceiver Resampler to map to linguistic space for the LLMs to understand.
When properly fed visual data, LLMs comprehend the visual realm and respond to instructions.
Therefore, the quality of visual cues extracted by the visual encoder directly impacts the LLMs' inference quality. There are three popular categories of visual encoders: fully supervised, text-guided weakly supervised, and self-supervised. The common fully supervised method~\cite{touvron2021training} involves pretraining on a well-annotated dataset like ImageNet.
Text-guided weakly supervised models pre-train on large image-text pairs using contrastive learning. A notable model, CLIP~\cite{radford2021learning}, excels in various tasks without specific training. Our method falls under self-supervised pre-training~\cite{caron2021emerging,he2022masked}.

\textbf{Difference with previous works.}
This work is most related to existing masked image modeling works.  Different from static masking patterns in existing works, the proposed method introduces evolved masks to consistently gain the model with low, mid, and high-level visual cues modeling capabilities. Those evolved masks can be easily generated thanks to the organized hierarchy representing multi-levels of image contents. As shown in experiments, the proposed method outperforms existing works by clear margins across different tasks. Our previous conference version~\cite{feng2023evolved} reveals that specific mask patterns, such as grid-like and large block masks, enhance the learning of distinct visual knowledge. We thus propose the evolved masking strategy by combining different mask patterns. However, simply combining mask patterns makes the proposed method inefficient and hard to tune due to considerable hyper-parameters. This updated version thus introduces more principled methodology by formulating the evolved masking as a hierarchical masking procedure that employs a unified mask generation process controlled by a single hyper-parameter, i.e., the masking depth. It enables a smooth transition between masking low-level and higher-level visual cues, leading to substantial performance enhancement. We also align the proposed method with the current research on LLMs and evaluate it across various tasks requiring different levels of visual cues, including Visual Question Answering, Object Counting, Multi-Class Identification, partial-duplicated retrieval, etc. Extensive evaluations further show the promising performance of the evolved masking strategy.

\section{Formulation}

Given an unlabeled image dataset $\mathcal{D}=\{I\}$, the goal of masking image modeling is to endow pre-trained model with the ability to extract distinguishable features from visual signals.
For an input image $I\in \mathbb{R}^{HW \times 3}$, where $H, W$ are the spatial size, we generate a binary mask $\mathcal{M} \in \{0,1\}^{HW}$ on $I$, and apply $\mathcal{M}$ to self-supervised learning.
Denote by $0<r<1$ the masked ratio, controlling number of masked patches, that
\begin{equation}
  \sum_{i=0}^{HW} \mathcal{M}_i = HW \times r.
\end{equation}
Specifically, we adopt an encoder-decoder structure with learnable parameters $\theta, \theta^\prime$ to recover $I$ from a masked input. For the $t$-th training epoch, the training objective is denoted as,
\begin{equation}~\label{eq:objective}
  \arg \min_{\theta, \theta^\prime} \mathop{\mathbb{E}}\limits_{I \sim \mathcal{D}} \mathcal{H} (\operatorname{G}_{\theta^\prime}^{(t)}(\operatorname{F}_\theta^{(t)}(I \odot {\mathcal{M}})), I \odot (1-{\mathcal{M}})),
\end{equation}
where $\odot$ is the element-wise product, $I \odot \mathcal{M}$ denotes the masked input. $\operatorname{F}_\theta^{(t)}$ and $\operatorname{G}_{\theta^\prime}^{(t)}$ are encoder and decoder in $t$-th training epoch, respectively. $\mathcal{H}(\cdot, \cdot)$ is the similarity measurement, \emph{e.g.,} $l2$-distance~\cite{he2022masked} or cross-entropy~\cite{bao2021beit}.

Masks $\mathcal{M}$ largely determine encoded cues in the optimized parameters $\theta$ and $\theta^\prime$ according to Eq~\eqref{eq:objective}.
In particular, there is a multitude of statistical dependencies among pixels, determined by their spatial location and visual semantics~\cite{anandkumar2013learning,huang2022latent,kong2023understanding}.
The network learns connections between underlying visual cues depending on masked and visible contents.
Visual modeling capability of a well-pretrained neural network allows it to transform
raw pixels into various levels of abstract concepts for diverse
tasks.

\begin{figure}
  \centering
  \includegraphics[width=1\linewidth]{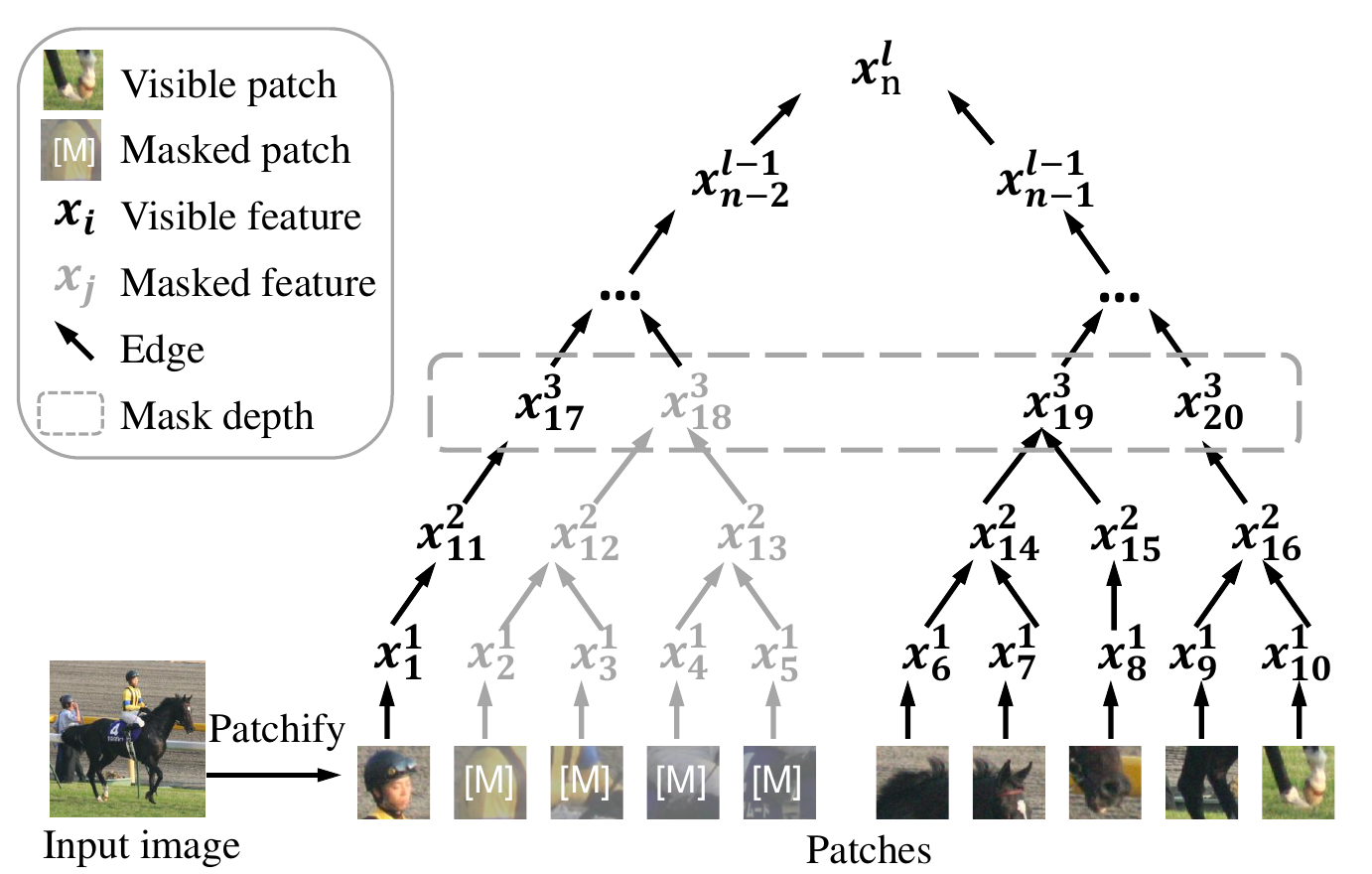}
  \caption{Illustration of established visual cue hierarchy structure $\mathcal{T}$. The MIM is performed by reconstructing masked patches according to visible regions. Different mask patterns can be adopted by alternating the mask depth on the hierarchy structure, e.g., masking nodes on the 3$^{rd}$ level or the 1$^{st}$ level. }
  \label{fig:hiear_formulatoin}
\end{figure}

To facilitate mask generation, we establish a binary tree structure, $\mathcal{T}:=(V, E)$, to represent the relationships between different visual cues, as illustrated in Fig.~\ref{fig:hiear_formulatoin}.
It consists of a set of latent nodes $V:=\{x_i^a\}$, where subscript  $i$ represents the sequence index, superscript $a$ denotes layer index. $E:=\{e_{i,j}\}$ denotes a set of directed edges, where $e_{i,j}$ connects nodes $x_i^a$ and $x_j^b$, demonstrating a causal relationship between them. It also indicates the visual cue dependencies the network learned. To simplify the presentation, we further define the set of parent and children of $x_i^a$ as $Pa_i^a$ and $Ch_i^a$. We also use $De_i^a$ to denote descendant nodes of node $x_i^a$.
Leaf nodes $x_i^1$ with superscript $a=1$ correspond to image patches, where $|\{x_i^1\}|= H \times W$ is the number of image patches.
Variables $x_i^a$ in lower layers correspond to intricate details such as texture. Latent variable $x_i^a$ with $a>$1 carries more complicated semantics. As illustrated in Fig.~\ref{fig:hiear_formulatoin}, nodes $x_{6}^1$ and $x_{7}^1$ depict certain details in the head of horse and node $x_{8}^1$ portrays the mouth. The node $x_{19}^3$ represents the entire horse head.
%


Constructing the $\mathcal{T}$ requires to model relationship among image patches. MIM pre-training is implemented by inferring masked cues from visible patches. This procedure supervises the network to learn relationships between different regions, hence developing certain visual cue modeling ability as illustrated in previous image representation~\cite{hinton2022represent,sabour2017dynamic} and generation~\cite{maaloe2019biva,vahdat2020nvae,zhao2017learning} works. We hence leverage the visual modeling capability of the pre-trained network to parse input images and construct $\mathcal{T}$.
Specifically, we propose an Adaptive Hierarchy Establishment (AHE) module to leverage features extracted by the encoder $\operatorname{F}_\theta^{(t)}$. This can be conceptually denoted as,
\begin{equation}
  \mathcal{T}^{(t)}=\operatorname{AHE}(\operatorname{F}_\theta^{(t)}, I).
\end{equation}
Note that, it is easier to model low-level details than learning complicated semantics. We hence progressively construct the $\mathcal{T}$ from layer 1 to layer $l$ at different training epoches to fully leverage the model being pre-trained.

The constructed tree $\mathcal{T}^{(t)}$ is adopted to supervise the MIM at subsequent training epoches. As $\mathcal{T}$ represents visual cues at different levels, different mask patterns can be adopted by alternating masking depths.
As illustrated in Fig.~\ref{fig:hiear_formulatoin}, we can set masking depth to $a=3$, and generate masks by randomly selecting nodes $x^3_{18}$.
Descendant leaf nodes of $x^3_{18}$ will be masked. The network is hence optimized towards inferring the masked latent variable $x^3_{18}$ according to other visible patches. Setting a lower mask depth to $a=1$ would pose easier MIM tasks of modeling low-level details. This procedure enables the MIM task to adaptively generate mask patterns at different training stages.


We hence leverage $\mathcal{T}^{(t)}$ to evolve the mask patterns at different training stages to enhance the generalization capability to different visual cues. Simple masks are generated to learn low-level visual cues at the initial training stage. More complicated masks are hence generated to learn object semantics at a later training stage.
We use the training epoch $t$ to represent different training stages, and hence determine the mask depth with $t$. The mask is generated in the Evolved Mask Generation (EMG) module, which is denoted as
\begin{equation}
  \mathcal{M}^{(t)}=\operatorname{EMG}(\mathcal{T}^{(t)},t).
\end{equation}

\begin{figure}[t]
  \centering
   \includegraphics[width=1\linewidth]{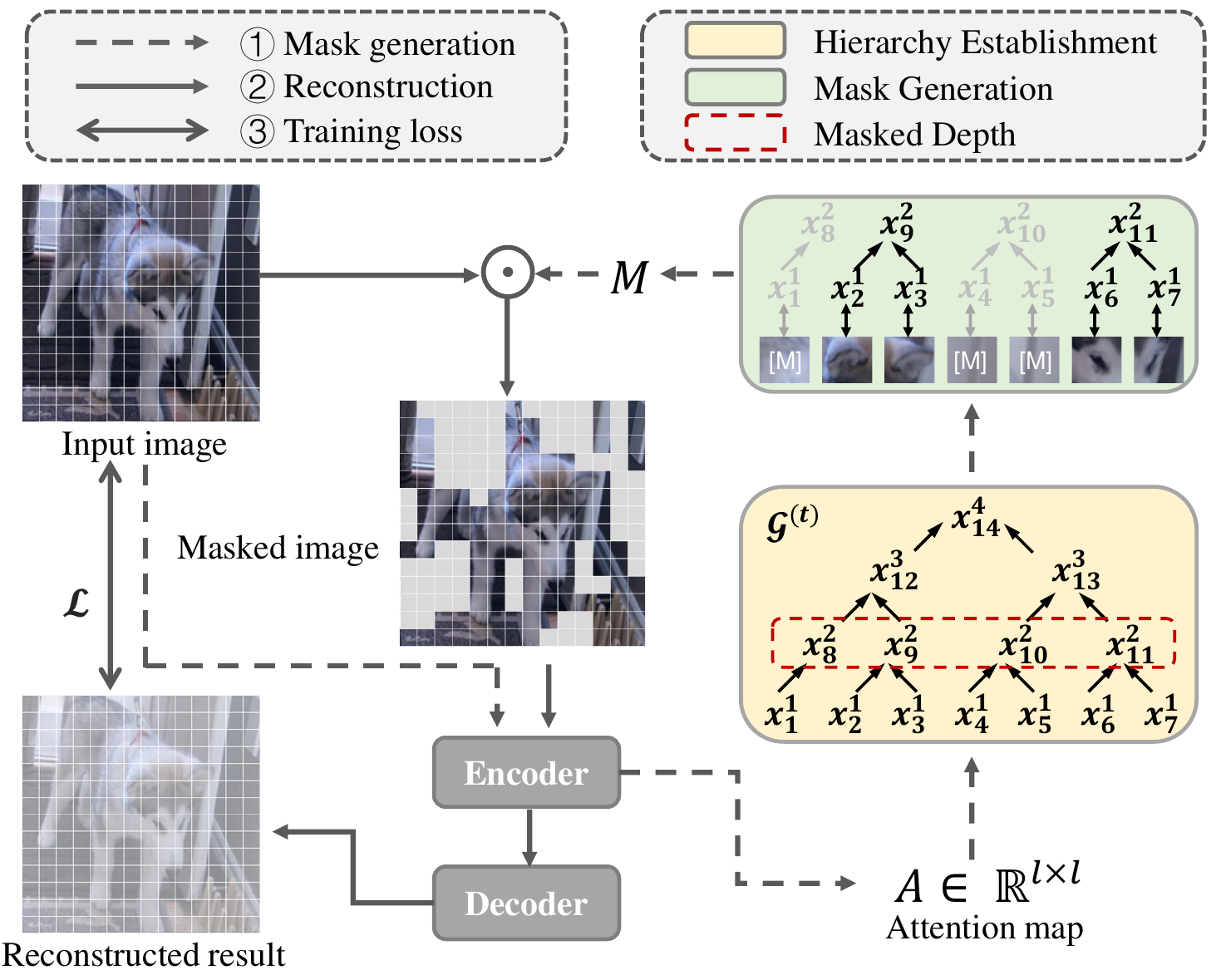}
   \caption{
    {The pipeline of proposed Evolved Hierarchical Masking} using MAE~\cite{he2022masked} as an example.
    Input image is fed into the encoder extracting the attention map $A$ to reflect the similarity between each pair of patches.
    The Adaptive Hierarchy Establishment module organizes leaf nodes and latent variables into the hierarchy $\mathcal{T}$.
    Based on $\mathcal{T}$, the Evolved Mask Generation module generates masks $\mathcal{M}$ on a specific masking depth.
   } \label{fig:method}
\end{figure}

The generated masks are hence adopted in the MIM task, and supervise the model pre-training with Eq.~\eqref{eq:objective}. 
Fig.~\ref{fig:method} illustrates the pipeline of the proposed method. The method starts by feeding the input image into encoder $F_\theta^{(t)}$ of the current epoch $t$. Similarities among image patches are then used to create the hierarchical structure $\mathcal{T}$ on the fly.
Masks $\mathcal{M}^{(t)}$ are generated by performing masking at a certain depth on $\mathcal{T}$, which evolves with pretraining epochs $t$.
The masked image is adopted for MIM, which is supervised by a loss between the original and recovered images for model optimization.
Our method iteratively performs the hierarchy establishment and mask generation, and finally leads to a more accurate hierarchy $\mathcal{T}$, and a more general pre-trained model.

The following part proceeds to present details in the Adaptive Hierarchy Establishment module and Evolved Mask Generation module, respectively. For brevity, we omit the superscript $(t)$ denoting the training epoch in the following section.

\section{Methodology}



\subsection{Adaptive Hierarchy Establishment}


Adaptive Hierarchy Establishment aims to organize image contents into a hierarchy reflecting their relationships. As the model being trained is gaining stronger visual contents modeling capability, we leverage it to compute similarity among image patches, and refer to the similarity cues for $\mathcal{T}:=(V, E)$ construction.
This adaptive hierarchy is dynamically constructed each time an image is input into the system.
In the early stages of training, the generated hierarchy is nearly random, and the masks are similarly random. The model learns visual knowledge from these random masks and generates a more accurate hierarchy, which in turn produces more challenging masks that are appropriate to the model`s current capabilities. This ensures that the model's capabilities and the difficulty of the masks are on par, fostering a kind of adversarial training mechanism that guides their co-evolution.


The hierarchy established needs to satisfy two requirements: (1) related patches are adjacent in the structure, and (2) it supports different levels of granularity. A tree structure is the most straightforward structure meets these criteria. In particular, a binary tree is efficient and computationally lightweight, as it only requires the calculation of pair-wise similarities. Hence, we use the binary tree as a simple representation of $\mathcal{T}$, and formalize its establishment as an optimal problem. 

Our goal is to organize the nodes into a binary tree to maximize the sum of similarity of adjacent sibling elements.
Using $\operatorname{S}(\cdot)$ as the similarity computation, the objective of $\mathcal{T}$ can be denoted as
\begin{equation}~\label{eq:objective-t}
  \mathcal{T} = \arg \max_{\mathcal{T}} \sum_{i=H\times W+1}^{n} \operatorname{S}(Ch_i^a),
\end{equation}
where $Ch_i^a$ is the set of children nodes for node $x^a_i$.

A series of works~\cite{bao2021beit,caron2021emerging} have demonstrated that transformer attention map can reflect the semantic relationship between tokens and a higher attention value represents a stronger relationship. We directly utilize the attention map from the current pre-trained model $\operatorname{F}_{\theta}$ to measure the similarity between the nodes without introducing extra parameters. This strategy also ensures the accuracy of generated $\mathcal{T}$ is on par with the modeling capacity of the trained $\operatorname{F}_{\theta}$.

Denoted as $A=\{A_{i,j}\} \in \mathbb{R}^{HW \times HW}$, the attention map produced by $\operatorname{F}_{\theta}$ provides similarity cues of leaf nodes in $\mathcal{T}$. For instance, $A_{i,j}$ reflects the relationship between the $i$-th and $j$-th patches, which are leaf nodes $x_i^1$ and $x_j^1$ in $\mathcal{T}$.
The similarity between two leaf nodes can be represented as
\begin{equation}
  \operatorname{S}(x_i^1,x_j^1)=A_{i,j}.    \\
\end{equation}
We hence use $A$ as a foundation for constructing $\mathcal{T}$ in a bottom-up way. $A$ doesn't provide similarity cues for internal nodes at layers larger than 1.
We measure their similarity by referring to their descendant leaf nodes. The similarity measurement can be computed as
\begin{equation}\label{eq:similarity}
  \operatorname{S}(x_i^a,x_j^b)=      \frac{\sum_{x_k^c \in Ch_i^a } \sum_{x_m^d \in Ch_j^b } \operatorname{S}(x_k^c,x_m^d)}{|Ch_i^a||Ch_j^b|},
\end{equation}
where $|Ch|$ denotes the number of children nodes, and we define the number of children of leaf nodes as 1.

Based on the computed similarity cues of $\operatorname{S}(\cdot)$, the training objective in Eq.~\eqref{eq:objective-t} can be achieved by iteratively merging closely related nodes into a parent node. It starts from merging leaf nodes, and continues until the root node is generated. 
The initial edge set $E_0$ is empty and the initial node set $V_0$ only contains leaf nodes with cardinality $H\times W$, i.e.,
\begin{equation}
  \begin{aligned}
    &  V_0=\{x_1^1,x_2^1,\ldots,x_{H\times W}^1\}. \\
    \end{aligned}
\end{equation}
At the $k$-th iteration, the most relevant nodes $x_i^a$ and $x_j^b$ are selected from $V_k$ to form a new internal node as their parent, which can be written as
\begin{equation}
  x_k^c = \{x_i^a, x_j^b\}=\arg \max_{(x_i^a, x_j^b)} \operatorname{S}(x_i^a, x_j^b),  x_i^a, x_j^b \in V_k.
\end{equation}
The hierarchical level of $x_k^c$ is determined based on the  level of its child nodes, which is
\begin{equation}
  c={max(a,b)+1}.
\end{equation}
The new node $x_k^c$ and new edges are added to $\mathcal{T}$. Meanwhile, its children nodes are removed from $\mathcal{T}$, i.e.,
\begin{equation}
    E_{k+1}=E_{k} \cup \{e_{a,k},e_{b,k}\}, \\
\end{equation}
\begin{equation}
Pa_i^a=Pa_j^b=x_k^c,
    Ch_k^c=\{x_i^a, x_j^b\}, \\
    \end{equation}
\begin{equation}\label{eq:merge}
  V_{k+1}=(V_k \, \backslash \,  \{x_i^a, x_j^b\}) \cup {x_k^c}.
\end{equation}

The above procedure continues for $H\times W-1$ iterations and ends with a hierarchical structure $\mathcal{T}$. Note that, Eq.~\eqref{eq:similarity} refers to leaf nodes to compute the similarity between internal nodes. We hence do not need to compute the feature representation of each internal node. Storing node similarities into a matrix avoids recursive similarity calculation as in Eq.\eqref{eq:similarity}, thus accelerates the computation. There are other ways to accelerate the $\mathcal{T}$ generation. Since the iterations are repeated for a constant $H\times W-1$ times for every image, this procedure can be efficiently batch parallelized. The computational time complexity of hierarchy establishment is $\mathbf{O}(HW)$. $HW$ is typically a constant 196, which constitutes only a small fraction of the total pretraining computational complexity.

\subsection{Evolved Mask Generation}

The Evolved Mask Generation module leverages the established $\mathcal{T}$ to generate mask $\mathcal{M} \in \{0,1\}^{H\times W}$ by referring to the current training epoch $t$.
Note that, the training epoch $t$ indicates the quality of established $\mathcal{T}$. For instance, a $\mathcal{T}$ at the early training stage with a small $t$ is not discriminative enough to show high-level semantics. It is hence adopted to generate simple mask patterns to reinforce the learning of low-level details in MIM. $\mathcal{T}$ established with a larger $t$ is hence adopted to supervise the learning of more complicated semantics.

To this end, we randomly select a portion of nodes at a specific depth $h$ on $\mathcal{T}$ according to $t$.
Masks are generated by removing image patches corresponding to selected nodes.
As the training progresses, $h$ gradually moves to the upper layers of $\mathcal{T}$ to evolve to more complicated mask patterns. 
$h$ is computed based on the ratio of the current  training epoch $t$ to the total number of epochs, i.e.,
\begin{equation}
  h=1+\lfloor{l \times \frac{t}{total\_epoches}}\rfloor,
\end{equation}
where $l$ denotes the height of the binary tree and $total\_epoches$ denotes the total number of pretraining epochs. This equation formulates the masking location $h$ as a parameter evolves linearly from shallower to deeper layers as the pretraining progresses. It simplifies the setting of hyperparameters and works well across various tasks on different datasets in Section \ref{sec:exp}.

Specifically, for nodes at the $h$-th layer of $\mathcal{T}$, we randomly select a subset of nodes, denoted as $\{x^h_i,x^h_{j},\cdots\}$.
This selection ensures that the total count of leaf nodes corresponding to these nodes equals $HW\times r$, where $r$ is the mask ratio.
We denote the set of descendant leaf nodes of these selected nodes as $V^{mask}$, defined as: 
\begin{equation}
  \begin{aligned}
    V^{mask}=\{De^h_{i}, De^h_{j}, \cdots\} \\
  \end{aligned}
\end{equation}
where nodes are randomly selected from the $h$-th depth of established tree $\mathcal{T}$, with the condition, that,
\begin{equation}
  |V^{mask}| = HW \times r
\end{equation}
If the total number of leaf nodes exceeds $HW \times r$, we truncate $V^{mask}$ to retain only the first $HW \times r$ nodes.



Finally, masks are generated by removing image patches corresponding to these selected descendant leaf nodes, that,
\begin{equation}
  \mathcal{M}_i=\left\{
    \begin{aligned}
    0, & \mbox{ if } x_i^1 \in {V^{mask}}   \\
    1, & \mbox{  otherwise. } \\
    \end{aligned}
    \right.
\end{equation}
This approach ensures consistent operations across images in a batch, which makes mask generation parallelizable and enhances efficiency.

During pretraining, we elevate the masking depth to higher layers.
In this way, the generated masks are adaptive to the current model parameters and evolve with the training epoch, gradually driving the network to learn from low-level details to more complex visual cue modeling.
The method hence provides a progressive and more reasonable learning difficulty curve, enhancing training efficiency and yielding robust pre-trained models for a variety of downstream tasks.

\subsection{Visualization}

\begin{figure}[t]
  \centering
   \includegraphics[width=1.0\linewidth]{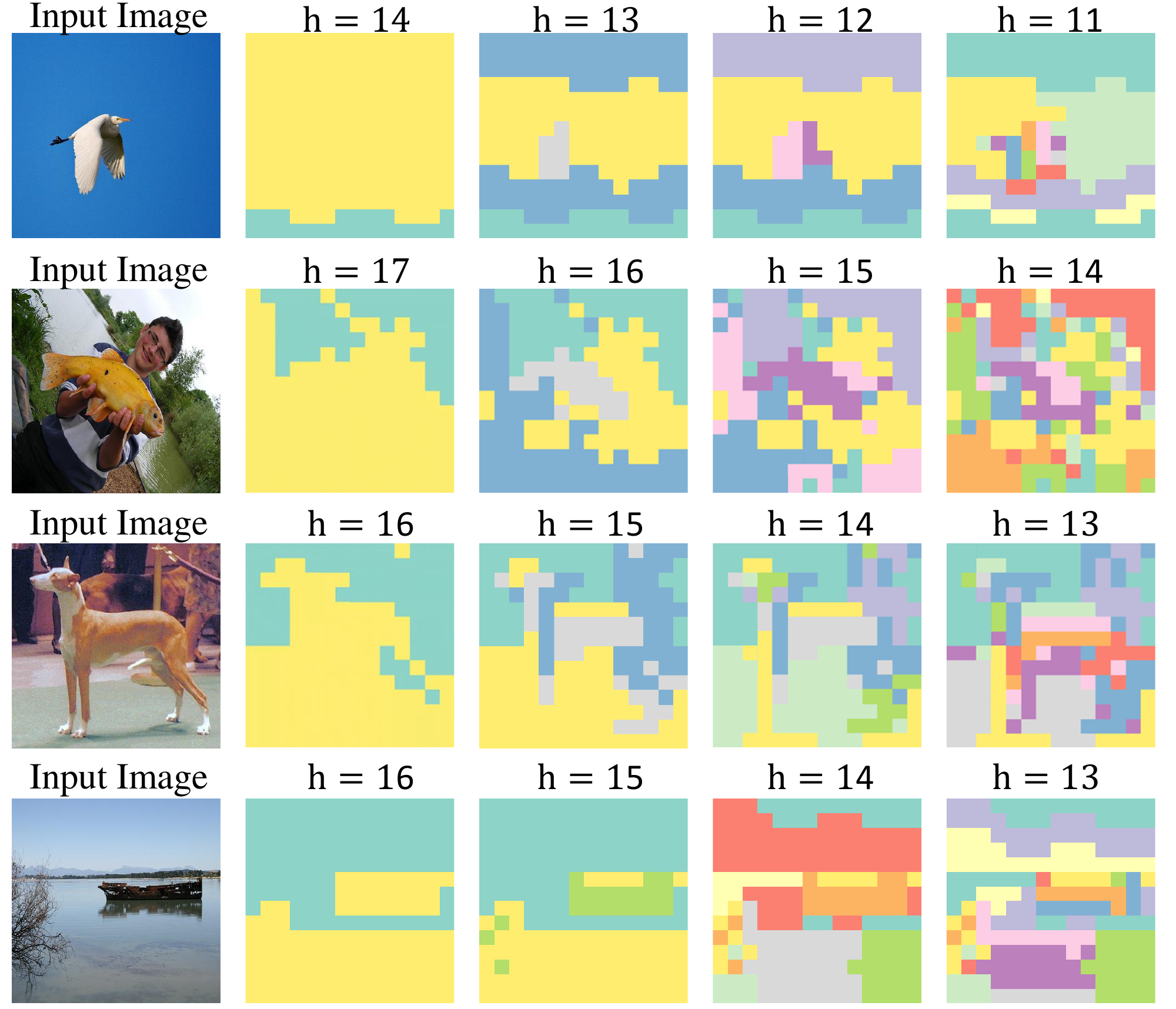}

   \caption{Visualization of image partition at different depth $h$ with the model after pretraining.
   }
   \label{fig:parti_visual}
\end{figure}

\textbf{Established tree structure.}
In Fig.~\ref{fig:parti_visual}, we visualize the partitioning results of an image at varying depths $h$. It can be seen that as the partition height $h$ decreases, the contents of divided partitions gradually shift from abstract concepts to specific, local parts. It's worth noting that our image partition differs from that of image segmentation: the former aims to identify relevant patches, while the latter focuses on pixel-level object division.
 For instance, in the first column of the figure, our method effectively divides the bird's body, a more complex region compared to the sky, into more parts.
On the other hand, given the varying likelihood of the sky appearing in different parts of the image, the incorporation of positional embedding leads to a top-to-bottom division of the sky into distinct regions for masking.

\begin{figure}[t]
  \centering
   \includegraphics[width=1.0\linewidth]{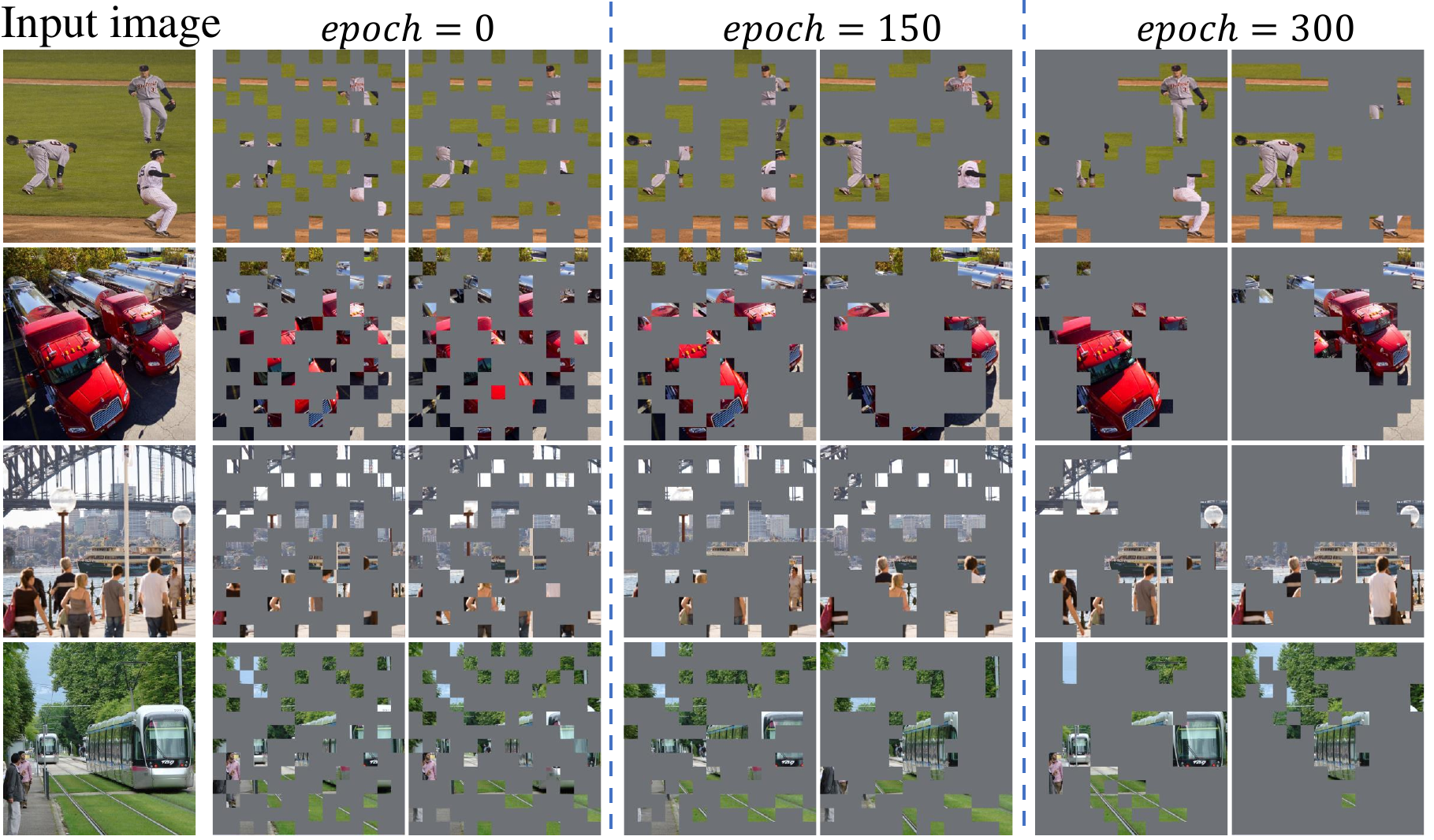}

   \caption{{Visualization of generated masks} in distinct training stages.
The shown masks come from using the MAE~\cite{he2022masked} method with ViT-Base~\cite{dosovitskiy2020image} as the backbone, trained for 300 epochs.
   }
   \label{fig:Visualization}
\end{figure}

\textbf{Generated masks.}
Guided by the hierarchy structure, the generated mask contains visual hints of different abstraction levels. This guides the network to learn progressively from simple to complex, and from specific textures to general meanings.
The masks also ensure a smooth difficulty curve for the learning process.
Fig.~\ref{fig:Visualization} visualizes the masks generated by the Evolve Hierarchy Masking in various pre-training stages.
In the initial pre-training phase,  the generated mask provides visual hints for most components, aiding the network in recognizing these components through texture details.
This simple task helps the model to learn essential low-level visual cues and facilitates rapid convergence.
As training advances,  the masked patches drop more complete objects, pushing the network to build associations within and among high-level semantics.
The generated masks are also more diverse, making recovery challenging while enhancing the learning of a broader range of visual knowledge.
For example, in the first image of Fig.~\ref{fig:Visualization},  the initial mask preserves visual hints for all three baseball players, grass, and playground.
The network can easily infer missing pixels using adjacent similar components.
After 300 training epochs, the mask largely removes some baseball players, necessitating the network to rely on the learned visual hierarchy to restore it.
This enhances the capability of grasping visual cues, providing stronger initial parameters for downstream tasks.

\begin{figure}[t]
  \centering
  \small
   \includegraphics[width=1\linewidth]{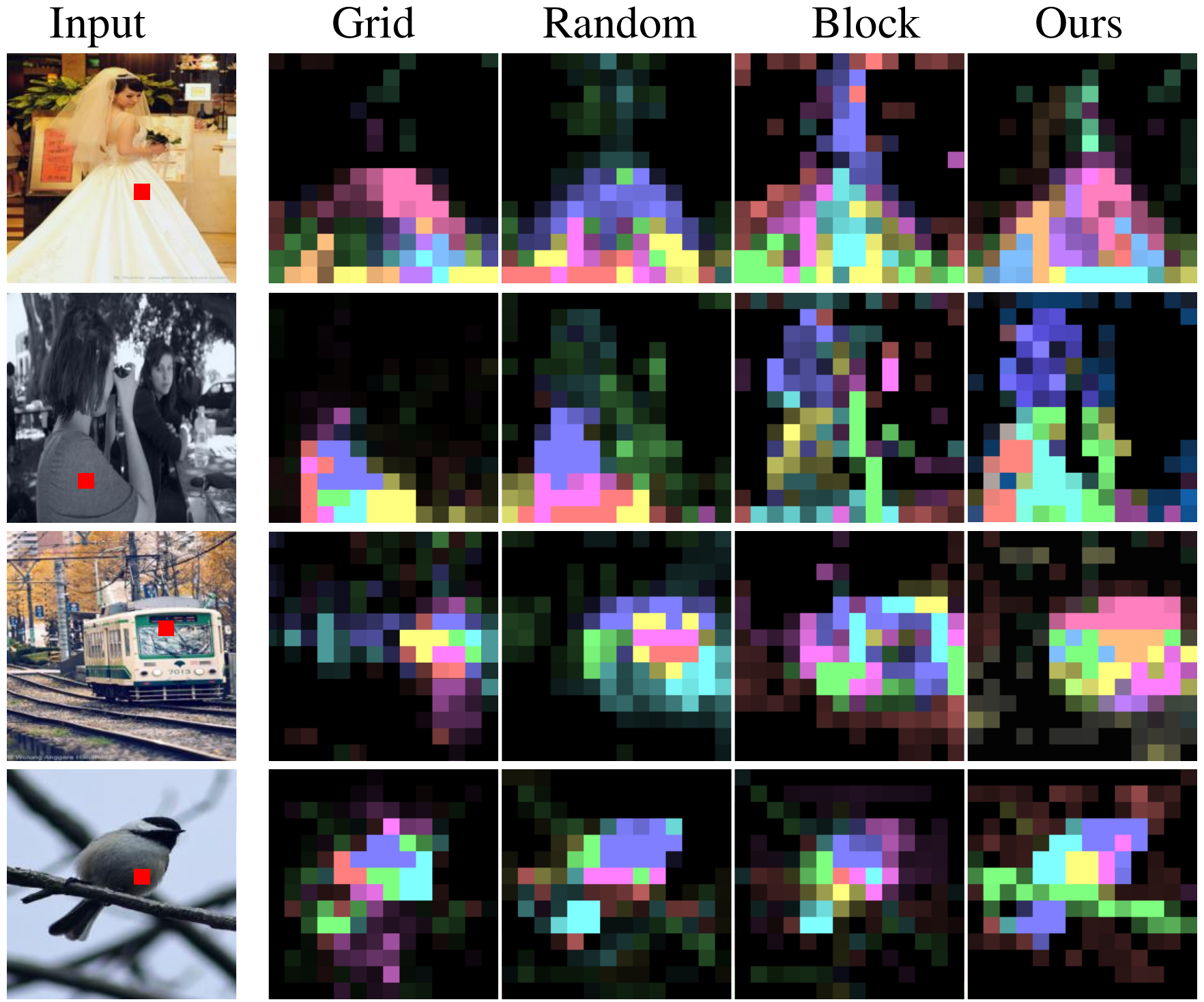}

   \caption{\textbf{Visualization of attention map} for sampled patches under different masking methods. A red dot in the input image annotates the sampled patch. Attention heads are encoded in various colours, and the brightness indicates the attention value. }
   \label{fig:attn_map}
\end{figure}

\textbf{Attention map of the pre-trained model.}
In Fig.\ref{fig:intro_curve} (b), we show the size of the attended area, where these methods exhibit distinct sizes.
To further explore the reasons, we visualize the attention maps of some patches in Fig.~\ref{fig:attn_map}.
It can be seen that grid-wise masking makes the pre-trained model focus on adjacent texture-similar features rather than semantically-similar ones.
In the first image shown in Fig.~\ref{fig:attn_map}, for grid-wise masking, the token on the wedding dress only focuses on the clothes patches, not the person wearing it.
Since there are always nearby visible patches to provide cues, the model tends to reconstruct the masked content based on these visible patches, which makes it attend mainly to nearby resemble patches.
Yet, the other masking approaches, to varying degrees, allow the network to learn more about high-level relationships.
When dealing with masks covering objects with fewer visible patches, the model must learn to model the relationship between global cues to predict these objects effectively.
Random masks play a role in compensating for the limitations of grid masks in terms of learning high-level semantics. It helps capture a bit more connection between objects. In comparison, block-wise masks enable the model to establish larger-scale connections between objects, though less accurately.

\section{Experiments}\label{sec:exp}

We evaluate our approach on seven tasks, including Image Classification, Semantic Segmentation, Detection, Object Counting (OC), Multi-Class Identification (MCI), Visual Question Answering (VQA), and Landmark Retrieval.

\subsection{Datasets and Evaluation Metric}\label{sec:dataset}

\emph{ImageNet-1K}~\cite{deng2009imagenet} dataset is made up of approximately $1.28M$ images for training and $50K$ images for validation with 1000 categories. Following literature~\cite {he2022masked,bao2021beit,xie2022simmim}, we conduct model pre-training on ImageNet-1K training images without utilizing labels.
Top-1 accuracy of finetuning and linear probing are reported for classification performance.

\emph{ADE20K}~\cite{zhou2017scene} is a challenging scene parsing benchmark. with 150 detailed semantic concepts. The training set includes 20,210 images with 150 semantic classes. The validation and test set contain 2,000 and 3,352 images respectively. In accordance with the standard evaluation method\cite{zhou2017scene}, we present the average mean Intersection-over-Union (mIoU) and mean Accuracy (mAcc) across all classes.

\emph{MS-COCO}~\cite{lin2014microsoft}, an abbreviation for Microsoft Common Objects in Context, is a large-scale object detection, segmentation, and captioning dataset.
It is comprised of more than $330K$ images and $1.5M$  object instances. It has 91 common objects found in their natural context which a 4-year-old human can easily recognize.
For detection performance, we report bounding box average precision (AP-box).
We report Top-1 accuracy on MS-COCO for both object counting and multi-class identification tasks.

\emph{VCR}~\cite{zellers2019vcr}, an abbreviation for  Visual Commonsense Reasoning, is made up of more than $110k$ images for cognition-level visual understanding. Following \cite{wang2023makes}, we construct the MCI task by asking the model if a certain object exists in the image with the prompt {\it``Question: Does {obj} exist in the image? Answer:''.}
The model is expected to answer  {\it``Yes/No''}, resulting in a binary classification problem.
We take out the "yes" or "no" from the initial sentence the model generates and match it with the correct answer.
Accuracy are reported for this task.

\emph{VQAv2}~\cite{goyal2017making} is a large-scale multi-modality dataset built on the top of VQA dataset~\cite{antol2015vqa}. It contains approximately 1.1 Million (image, question) pairs, comprising of $ $$\sim$$900K$ abstract scene picture and $ $$\sim$$204K$ real images from COCO with approximately 13 Million associated answers. The real images dataset contains $ $$\sim$$614K$ free-form natural language questions (3 questions per image), and over 6 million free-form but concise answers (10 answers per question).
Modern Language Models (MLLMs) generally produce more than one sentence. This makes it challenging to directly assess them using the standard evaluation protocol, which necessitates an exact match between predictions and ground truth. To address this, we follow \cite{wang2023makes} and modify the original evaluation protocol. We consider the first sentence generated by the MLLM as the prediction and deem it accurate if it includes the correct answer from the ground truth.

\emph{Oxford}~\cite{philbin2007object} and \emph{Paris}~\cite{philbin2008lost} are used to evaluate our model on landmark retrieval task, referred to as $\mathcal{R}$Oxf and $\mathcal{R}$Par, respectively. Each dataset has 70 queries and includes 4993 and 6322 database images, respectively. Performance is measured using mean average precision (mAP). Besides, the pre-trained models are finetuned for landmark retrieval tasks on clean Google Landmarks dataset v2 (GLDv2)~\cite{weyand2020GLDv2} following \cite{cao2020unifying}. The clean version of GLDv2 contains 1.58M images with 81K classes.

Following GVT~\cite{wang2023makes}, the object counting task is constructed on MS-COCO~\cite{lin2014microsoft} dataset by instructing the model to enumerate instances of a specific object present within an image, using the prompt {\it"Question: How many {obj} are there in the image? Answer:"}. The number in the first generated sentence is extracted as the prediction and is compared with the ground truth number.
This task is treated as a classification assignment, and the performance is measured by the model's top-1 prediction accuracy.

\subsection{Implementation Details} \label{sec:implementation}

We use different capacity Vision transformers as the backbones in our study, \emph{i.e.,} ViT-S, ViT-B~\cite{dosovitskiy2020image} and Swin transformer~\cite{liu2021swin}.
We apply the proposed masking method to three popular MIM models, \emph{i.e.,} MAE~\cite{he2022masked}, BEiT~\cite{bao2021beit} and SimMIM~\cite{xie2022simmim}.
Both pre-training and fine-tuning for classification are conducted on ImageNet-1K~\cite{deng2009imagenet} dataset under $224 \times 224$ resolution.
The images are patchifeid with patch size $p=16$ for ViT and $p=4$ for the Swin transformer.
Models structure and optimization settings follow that in the corresponding works.
Pretrained models are optimized by the AdamW~\cite{loshchilov2017decoupled} optimizer and we set $\beta_1$ to 0.9 and $\beta_2$ to 0.95.
By default, the models are pretrained for 300 epochs for a fair comparison, the mask ratio $r$ is set to $0.75$ following the original model~\cite{he2022masked}.

After pre-training, the model parameters are utilized as initializations for various downstream tasks.
For linear probing, we follow the standard setting in MAE~\cite{he2022masked}, disabling many common regularization strategies, including mixup, cutmix, drop path and setting weight decay as zero.
For semantic segmentation, we take UpperNet~\cite{xiao2018unified} as the framework and use the pre-trained encoder parameter to initialize the model backbone.
The segmentation models are then fine-tuned on ADE20K for $160K$ iterations with the default setting.
The input resolution is $512 \times 512$ without using multi-scale test.
For object detection, we adapt the pretrained transformer to take the place of the vanilla  FPN backbone~\cite{lin2017feature} in Mask R-CNN~\cite{he2017mask} following \cite{he2022masked} and fine-tune the model for $30$ epochs.
The ViT modules are split into 4 subsets, each performing convolutions to adjust the feature maps' sizes. This creates various scales: stride 4, 8, 16, and 32, similar to a standard ResNet.
We use the AdamW~\cite{loshchilov2017decoupled} optimizer with a weight decay of 0.05.
The standard $1 \times$  and $3 \times$ schedules are applied, where initial learning rate is $3 \times 10^{-4}$ and it decays by a factor of 10 after $3/4$ and $11/12$ of finetuning epochs.
For landmark retrieval, we follow DELG~\cite{cao2020unifying} and replace the visual backbone with the pre-trained models. These models are fine-tuned for 6M steps, 
optimizing with SGD and a momentum of 0.9, 
and a learning rate that decreases linearly to zero after reaching the desired step count.
Our method supports multiple hardware platforms and already
supports training and deployment on the Ascend 910B NPU.

We further evaluate the ability of the proposed pre-trained models to understand visuals based on semantic understanding and fine-grained detail perception, following GVT~\cite{wang2023makes}.
This assessment involves tasks involved include visual question answering, multi-class identification and object counting. They focus on evaluating high-level semantics, mid-level visual cues, and low-level details, respectively.
We use different pre-trained visual encoders to represent visual contents as token sequences. These sequences are then aligned with the input space of the Large Language Model (LLM) using the Perceiver Resampler~\cite{jaegle2021perceiver}.
The aligned features are subsequently fed into the LLM for final prediction.
The ensemble of models is fine-tuned jointly, or only the Perceiver Resampler~\cite{jaegle2021perceiver} is fine-tuned while keeping the pre-trained language model fixed.
We resize the input images to $224 \times 224$ and apply basic data augmentations: random resized crop and horizontal flipping. The model undergoes fine-tuning for $20K$ iterations using the AdamW~\cite{loshchilov2017decoupled} optimizer with a learning rate of $1 \times 10^{-6}$.

\subsection{Ablation Study} \label{sec:ablation}
Models in ablation experiments are built upon the ViT-S backbone~\cite{dosovitskiy2020image} and asymmetric MAE architecture~\cite{he2022masked}.

\begin{table*}[t]
  \centering
  \small
  \caption{Downstream tasks performance after fine-tuning. Models are pre-trained on imageNet-1K~\cite{deng2009imagenet} with different masking methods. We report imageNet-1K Top-1 accuracy, ADE20K mIoU~\cite{zhou2017scene}, and COCO AP-box~\cite{lin2014microsoft} for classification, semantic segmentation and object detection, respectively.
  The first line is the performance with random initialization in fine-tuning.
  The best performances for each task are noted with a grey back colour.}
  \label{tab:mask_method_downstream}

  \begin{tabular}{c|cccc|cccc|cccc}
     \hline
    {\color[HTML]{000000} } &
      \multicolumn{4}{c|}{{\color[HTML]{000000} Classification}} &
      \multicolumn{4}{c|}{{\color[HTML]{000000} Segmentation}} &
      \multicolumn{4}{c}{{\color[HTML]{000000} Detection}} \\
    \multirow{-2}{*}{{\color[HTML]{000000} Epochs}} &
      {\color[HTML]{000000} grid} &
      {\color[HTML]{000000} random} &
      {\color[HTML]{000000} block} &
      {\color[HTML]{000000} ours} &
      {\color[HTML]{000000} grid} &
      {\color[HTML]{000000} random} &
      {\color[HTML]{000000} block} &
      {\color[HTML]{000000} ours} &
      {\color[HTML]{000000} grid} &
      {\color[HTML]{000000} random} &
      {\color[HTML]{000000} block} &
      {\color[HTML]{000000} ours}   \\ \hline
    \multicolumn{1}{c|}{{\color[HTML]{000000} Random ini.}} &
      {\color[HTML]{000000} 71.53} &
      {\color[HTML]{000000} 71.53} &
      {\color[HTML]{000000} 71.53} &
      {\color[HTML]{000000} 71.53} &
      \multicolumn{1}{c}{{\color[HTML]{000000} 21.17}} &
      \multicolumn{1}{c}{{\color[HTML]{000000} 21.17}} &
      \multicolumn{1}{c}{{\color[HTML]{000000} 21.17}} &
      \multicolumn{1}{c|}{{\color[HTML]{000000} 21.17}} &
      {\color[HTML]{000000} 19.31} &
      {\color[HTML]{000000} 19.31} &
      {\color[HTML]{000000} 19.31} &
      {\color[HTML]{000000} 19.31} \\ \hline
    {\color[HTML]{000000} 100} &
      {\color[HTML]{000000} 78.28} &
      {\color[HTML]{000000} 78.11} &
      {\color[HTML]{000000} 77.85} &
      \cellcolor[HTML]{EFEFEF}{\color[HTML]{000000} 78.29} &
      {\color[HTML]{000000} 36.65} &
      {\color[HTML]{000000} 37.60} &
      {\color[HTML]{000000} 36.02} &
      \cellcolor[HTML]{EFEFEF}{\color[HTML]{000000} 37.82} &
      \cellcolor[HTML]{EFEFEF}{\color[HTML]{000000}  36.11 } &
      {\color[HTML]{000000} 32.70 } &
      {\color[HTML]{000000} 31.78} &
      {\color[HTML]{000000} 34.62}  \\
    {\color[HTML]{000000} 200} &
      {\color[HTML]{000000} 78.63} &
      {\color[HTML]{000000} 79.20} &
      {\color[HTML]{000000} 78.55} &
      \cellcolor[HTML]{EFEFEF}{\color[HTML]{000000} 79.92} &
      {\color[HTML]{000000} 36.67} &
      {\color[HTML]{000000} 38.81} &
      {\color[HTML]{000000} 36.65} &
      \cellcolor[HTML]{EFEFEF}{\color[HTML]{000000} 39.78} &
      {\color[HTML]{000000} 36.21} &
      {\color[HTML]{000000} 34.42 } &
      {\color[HTML]{000000} 33.15} &
      \cellcolor[HTML]{EFEFEF}{\color[HTML]{000000} 36.51} \\
    {\color[HTML]{000000} 400} &
      {\color[HTML]{000000} 79.11} &
      {\color[HTML]{000000} 79.42} &
      {\color[HTML]{000000} 79.29} &
      \cellcolor[HTML]{EFEFEF}{\color[HTML]{000000} 80.41} &
      {\color[HTML]{000000} 36.78} &
      {\color[HTML]{000000} 39.43} &
      {\color[HTML]{000000} 40.67} &
      \cellcolor[HTML]{EFEFEF}{\color[HTML]{000000} 41.37} &
      {\color[HTML]{000000} 35.19} &
      {\color[HTML]{000000} 35.21} &
      {\color[HTML]{000000} 35.08} &
      \cellcolor[HTML]{EFEFEF}{\color[HTML]{000000} 37.45} \\
    {\color[HTML]{000000} 800} &
      {\color[HTML]{000000} 79.34} &
      {\color[HTML]{000000} 79.77} &
      {\color[HTML]{000000} 79.69} &
      \cellcolor[HTML]{EFEFEF}{\color[HTML]{000000} 80.72} &
      {\color[HTML]{000000} 36.54} &
      {\color[HTML]{000000} 39.31} &
      {\color[HTML]{000000} 41.81} &
      \cellcolor[HTML]{EFEFEF}{\color[HTML]{000000} 42.02} &
      {\color[HTML]{000000} 34.62} &
      {\color[HTML]{000000} 38.73} &
      {\color[HTML]{000000}38.91      } &
      \cellcolor[HTML]{EFEFEF}{\color[HTML]{000000} 39.16} \\
      \hline
      Avg. & 78.84 & 79.13 & 78.85 & \cellcolor[HTML]{EFEFEF}{\color[HTML]{000000}79.84} & 36.66 & 38.79 & 38.79 & \cellcolor[HTML]{EFEFEF}{\color[HTML]{000000}40.25} & 35.53 & 35.27 & 34.73 & \cellcolor[HTML]{EFEFEF}{\color[HTML]{000000}36.94} \\
       \hline
    \end{tabular}
  \end{table*}

\textbf{Comparison with static masking.}
We first investigate the performance of static masking methods and the proposed evolved method on various downstream tasks in Tab.~\ref{tab:mask_method_downstream}.
It can be seen that different static masking methods exhibit distinct advantages.
For example, grid-wise masking gives the network better performance with fewer pre-training epochs. Compared with classification tasks, block-wise masking brings more performance improvements on dense prediction tasks under sufficient pre-training.
The properties exhibited by random masking lie in between, including the strengths and weaknesses of the two.
Meanwhile,  our method combines the advantages and overcomes the disadvantages by varying masking criteria along the pre-training process, which outperforms these static methods in both performance and efficiency.



\textbf{Mask ratio.}
\begin{table}[]
  \centering
  \small
  \caption{Comparison of imageNet-1K classification~\cite{deng2009imagenet} and ADE20K segmentation~\cite{zhou2017scene} with various mask ratios.
  We report Top-1 accuracy and mIoU respectively.
  } \label{tab:mask_ratio}
  \begin{tabular}{c|cccc} \hline
    \multirow{2}{*}{Mask   ratio} & \multicolumn{2}{c}{Classification} & \multicolumn{2}{c}{Segmentation} \\
         & Random & Ours  & Random & Ours  \\ \hline
    0.25 & 77.19  & 78.96 & 36.23   & 38.14 \\
    0.50 & 79.02  & 79.84 & 38.56  & 39.11 \\
    0.75 & 79.35  & 80.06 & 39.24  & 40.59 \\
    \hline
    \end{tabular}
  \end{table}
The process of restoring masked areas associates the visible pixels with the removed content.
This determines the visual dependencies learned by the pre-trained network, making an appropriate mask ratio crucial for pre-training. MAE~\cite{he2022masked} shows that for random masks, optimal ratios are surprisingly high. These ratios enable the network to fully grasp high-level meanings.
Tab.~\ref{tab:mask_ratio} shows the impact of various mask ratios on random masking and the proposed method.
It can be seen that the proposed method is more robust to mask ratio, outperforming random masking +1.77\% and 2.24\% for  ImageNet classification~\cite{deng2009imagenet} and ADE20K segmentation~\cite{zhou2017scene} at ratios of 0.25, respectively.
The discrepancy between the highest and lowest values is merely 1.1\% and 2.45\% for two tasks.
In comparison, the difference for random masking reaches 2.16\% and 3.01\% for classification and segmentation.
This is because, under a mask ratio of 0.25, random masking retains too much similar content, allowing the network to restore masked patches by extending lines or textures.
This issue becomes more evident in segmentation task.
The proposed method takes the difficulty of pre-text task into consideration.
Even at a low mask ratio, some components will be moved out as a whole, ensuring the network learns ample semantics.

\textbf{Hierarchical masking.}
\begin{table}[]
  \small
  \centering
  \caption{Comparison between hierarchical masking and basic masking patterns on ImageNet-1K top-1 ACC. and ADE20K mIoU. 
  } \label{tab:static_masks}
  \begin{tabular}{ccc}
    \hline
      Masking Method                     & Classification & Segmentation \\ \hline
  Large block                    & 77.89          & 37.16        \\
  Hierarchical (only deep)   & 78.76          & 39.01        \\ \hline
  Grid-like                      & 73.16          & 32.16        \\
  Hierarchical (only shallow) & 79.05          & 39.86       \\
  \hline
  \end{tabular}
  \end{table}
Some simpler masking methods, such as grid-like patterns for fine-grained masking and randomly large block masking for object-level masking, can also produce certain masking patterns. Tab.~\ref{tab:static_masks} compares simple masking patterns with the proposed method upon a ViT-S backbone, on ImageNet-1K classification and ADE20K segmentation tasks. We used randomly large block masking and our method with a deep masking depth to generate object-level masks. Similarly, we used grid-like masking and our method with a shallow masking depth for part-level masks. Note that we kept the depth of masking constant in our method, i.e., 20\% of the height for shallow masking, and 80\% for the deep one, to ensure a fair comparison. As shown in Tab.~\ref{tab:static_masks}, the proposed method achieves substantially better performance in both tasks than those simple masking methods, which demonstrates the effectiveness of hierarchical masking.

\textbf{Evolved masking depth.}
\begin{table}[t!]
  \centering
  \small
  \caption{
    Comparison between fixed masking depth and the evolved masking depth.  Models, built upon ViT-S, are pre-trained for 300 epochs and evaluated on  Oxford retrieval, ADE20K segmentation, and ImageNet-1K classification.
  }\label{tab:fixed_depth}
  \begin{tabular}{cccc} \hline
  Masking depth & Retrieval & Segmentation & Classification \\ \hline
  Only shallow  & 57.11     & 39.86        & 79.05          \\ 
  Only deep     & 55.07     & 39.01        & 78.76          \\ 
  Evolved       & 57.41     & 41.37        & 80.41         \\ \hline
  \end{tabular}
  \end{table}
  The concern arises whether a model, after undergoing training with evolved masking depths, may forget initially acquired low-level knowledge. To investigate this issue, we compare the effects of fixed and evolved masking depths in Tab.~\ref{tab:fixed_depth}.  The compared masking strategies include 1) only shallow: 20\% of the height as the shallow masking for low-level cues, 2) only deep: 80\% of the height as the deep masking for high-level semantics, and 3) evolved: the one that evolved with training. It can be seen that, compared to using only low-level masking, the evolved method achieves substantial performance improvements in classification and segmentation tasks that relatively prioritize higher-level semantics, i.e., +1.51\% and +1.36\%, respectively. It also achieves comparable performance in retrieval task that focus on low-level semantics (57.11\% vs. 57.41\%). The comparison between ``only deep" and ``only shallow" masking shows that low-level knowledge is indeed crucial for pre-training, enabling more efficient learning and faster convergence, especially in retrieval (57.11\% vs. 55.07\%). This demonstrates that high-level and low-level knowledge are not mutually exclusive or conflicting but rather, low-level knowledge is foundational for learning higher-level information. The evolved masking allows for learning high-level visual semantics on the basis of low-level perception, achieving better network performance in various downstream tasks.

  \begin{table}[]
    \centering
    \small
    \caption{Effect of more training epochs. Pretrained ViT-S models are validated on imageNet-1K Top-1 accuracy, ADE20K mIoU~\cite{zhou2017scene}, and COCO AP-box~\cite{lin2014microsoft}.
    }  \label{tab:extending_epoch}
    \begin{tabular}{c|ccc} \hline
    Epochs & Classification & Segmentation & Detection \\ \hline
    100    & 78.29          & 37.82        & 34.62     \\
    200    & 79.92          & 39.78        & 36.51     \\
    400    & 80.41          & 41.37        & 37.45     \\
    800    & 80.72          & 42.02        & 39.16     \\
    1600   & 81.41          & 42.79        & 40.85     \\
    2400   & 81.55          & 42.94        & 41.08  \\  \hline
    \end{tabular}
  \end{table}

\textbf{Training epochs.}
In our experiments, we selected 300 training epochs as the primary setting. 
Experiments in Table \ref{tab:extending_epoch} investigate extending training epochs on model performance. Generally, extending pretraining duration leads to enhanced performance, but it also increases computational costs. Notably, the current pretraining performance has not yet reached a plateau, suggesting potential for further improvement.

\begin{table*}[t]
  \centering
  \small
  \caption{Comparison of popular self-supervise learning methods on imageNet-1K~\cite{deng2009imagenet}.
  Evaluation protocols include top-1 linear probing accuracy and top-1 fine-tuning accuracy.
  All entries are on an image size of $224 \times 224.$
  }  \label{tab:sota}
  \begin{tabular}{llccccc} \toprule
  Method   & Pre-train data & Date                             & Extra model  & Epochs & Linear probing & Fine-tune \\ \hline
  \multicolumn{7}{l}{
    \cellcolor[HTML]{EFEFEF}\emph{Supervised traning   from scratch} }                                                    \\
  ViT~\cite{dosovitskiy2020image}      & IN-1K w/ label & ICLR 2021                        & -            & 300    & -      & 77.9     \\
  DeiT~\cite{touvron2021training}     & IN-1K w/ label & ICML   2021  & -            & 300    & -      & 81.8      \\ \hline
  \multicolumn{7}{l}{
    \cellcolor[HTML]{EFEFEF}\emph{Contrastive-based SSL   Pre-training}    }                                                       \\
  DINO~\cite{caron2021emerging}     & IN-1K          & CVPR 2021                        & momentum ViT & 1600   & 74.6   & 82.8      \\

  MoCo v3~\cite{chen2021empirical}  & IN-1K          & ICCV 2021                        & momentum ViT & 300    & \textbf{76.5}   & 83.2      \\
  MST~\cite{li2021mst}      & IN-1K          & NIPS 2021                        & MLP Head           & 100    & 75.0     & -         \\

  AttnMask~\cite{kakogeorgiou2022hide} & IN-1K          & ECCV 2022 & momentum ViT            & 100    & 76.1   & -         \\
   \hline
  \multicolumn{7}{l}{
    \cellcolor[HTML]{EFEFEF}\emph{MIM SSL   Pre-training}    }                                                       \\
    BEiT~\cite{bao2021beit}     & IN-1K+DALL-E   & ICCV 2021                        & dVAE~\cite{ramesh2021zero}         & 300    & 56.7   & 82.9     \\
  CAE~\cite{chen2022context}      & IN-1K          & arxiv 2022 & -            & 300    & 64.1   & 83.6      \\
  MAE~\cite{he2022masked}      & IN-1K          & CVPR 2022                        & -            & 300    & 64.4   & 83.6      \\
  SimMIM~\cite{xie2022simmim}   & IN-1K          & CVPR 2022                        & -            & 800    & 56.7   & 83.8      \\
  SemMAE~\cite{li2022semmae}   & IN-1K          & NIPS 2022 & iBOT~\cite{zhou2021ibot}         & 800    & 68.7   & 83.3     \\
  MFM~\cite{xie2022masked}      & IN-1K          & ICLR 2023 & -            & 300    & -      & 83.1      \\
  LocalMIM~\cite{wang2023masked}    & IN-1K    &   CVPR 2023     &   -   &   100   &  -  &  83.3  \\
  EPM~\cite{feng2023evolved}      & IN-1K          & CVPR 2023 & -            & 200    & 59.6      & 83.6      \\
  Ours     & IN-1K          & -                                & -            & 200    & 63.4 &  83.8     \\
  Ours     & IN-1K          & -                                & -            & 300    &  66.1  &  \textbf{84.9}    \\
  \bottomrule
  \end{tabular}
  \end{table*}

  \begin{table*}[]
    \centering
    \small
    \caption{Comparison of visual tokenizers of ViT-B~\cite{dosovitskiy2020image} with different pretraining methods. The tasks include comparing visual question answering, object counting, and multi-class classification on MSCOCO~\cite{lin2014microsoft}. Additionally, there's a multi-class classification on Visual Commonsense Reasoning~\cite{zellers2019vcr}. The best result is marked with bold.
    }  \label{tab:tokenizer}
    \begin{tabular}{ccccccccc} \toprule
      \begin{tabular}[c]{@{}c@{}}Joint\\ Tuning\end{tabular}                                  & Supervised & Visual Encoder & \begin{tabular}[c]{@{}c@{}}\# Pretraining\\ Images\end{tabular} & VQA COCO & OC COCO & \multicolumn{1}{c}{MCI COCO} & MCI VCR & Avg.   \\ \hline
    \multirow{6}{*}{\XSolidBrush}   & Fully      & DeiT~\cite{touvron2021training}           & 1.28M                 & 48.81   & 67.28   & 78.03    & 82.33   & 69.11 \\ \cline{2-9}
     & Weakly                & CLIP~\cite{radford2021learning}    & 400M  & \textbf{51.54} & 64.44          & 78.94                     & 83.75          & 69.67 \\ \cline{2-9}
     & \multirow{4}{*}{Self} & DINO~\cite{caron2021emerging}    & 1.28M & 49.40 & 66.23          & 73.55                     & 79.01          & 67.05 \\
     &                       & DINO v2~\cite{oquab2023dinov2} & 142M  & 50.56 & 67.91          & 78.63                     & 83.88          & 70.25 \\
     &                       & MAE~\cite{he2022masked}     & 1.28M & 48.21 & 67.15          & 76.46 & 81.80          & 68.60 \\
     &                       & Ours    & 1.28M & 49.75 & 67.42          & 78.85                     & 82.54          & 69.64 \\ \hline
    \multirow{6}{*}{\CheckmarkBold} & Fully      & DeiT~\cite{touvron2021training}           & 1.28M                 & 50.86   & 67.50   & 80.09                        & 83.82   & 70.57 \\ \cline{2-9}
     & Weakly                & CLIP~\cite{radford2021learning}    & 400M  & 47.87 & 66.71          & 80.98                     & 80.94          & 69.13 \\ \cline{2-9}
     & \multirow{4}{*}{Self} & DINO~\cite{caron2021emerging}    & 1.28M & 46.28 & 64.35          & 80.17                     & 82.90          & 68.43 \\
     &                       & DINO v2~\cite{oquab2023dinov2} & 142M  & 49.99 & 65.86          & 79.26                     & 82.62          & 69.43 \\
     &                       & MAE~\cite{he2022masked}     & 1.28M & 48.72 & 67.06          & 81.40                     & 84.08          & 70.19 \\
     &                       & Ours    & 1.28M & 50.94 & \textbf{68.36} & \textbf{81.79}            & \textbf{84.91} & \textbf{71.50} \\
     \bottomrule
    \end{tabular}
    \end{table*}

\subsection{Benchmark Performance}\label{sec:benchmark}

\textbf{Applying to various MIM methods.}
\begin{table}[t]
  \centering
  \small
  \caption{ImageNet-1K~\cite{deng2009imagenet} accuracy and ADE20K~\cite{zhou2017scene} performance of three popular MIM models before and after applying the proposed masking with fine-tuning.}
  \label{tab:baseline}
  \begin{tabular}{lcccc}
    \toprule
  \multicolumn{1}{c}{\multirow{2}{*}{Method}} & \multicolumn{2}{c}{Classification}                              & \multicolumn{2}{c}{Segmentation}                    \\ \cmidrule(r){2-3} \cmidrule(r){4-5}
  \multicolumn{1}{c}{}                        & \multicolumn{1}{c}{top-1. acc } & \multicolumn{1}{c}{top-5. acc} & \multicolumn{1}{c}{mIoU} & \multicolumn{1}{c}{aAcc} \\ \hline
  MAE~\cite{he2022masked}                       & 79.20 & 94.61 & 38.81 & 79.61 \\
  \multicolumn{1}{c}{+ours} & 80.06 & 95.10 & 40.23 & 79.87 \\ \hline
  BEiT~\cite{bao2021beit}                      & 79.05 & 94.57 & 38.39 & 79.31 \\
  \multicolumn{1}{c}{+ours} & 79.87 & 94.80 & 39.78 & 79.83 \\ \hline
  SimMIM~\cite{xie2022simmim}                    & 78.51 & 94.18 & 38.13 & 79.15 \\
  \multicolumn{1}{c}{+ours} & 79.21 & 94.79 & 39.45 & 79.73 \\
  \hline
  \end{tabular}
  \end{table}
We validate the effect of the proposed method on three popular MIM models, \emph{i.e.,} MAE~\cite{he2022masked}, BEiT~\cite{bao2021beit} and SimMIM~\cite{xie2022simmim} and evaluate their performance on imageNet-1K classification~\cite{deng2009imagenet} and ADE20K segmentation~\cite{zhou2017scene}.
For a fair comparison, the models are pre-trained using the official code for 200 epochs and fine-tuned on downstream tasks with consistent experimental settings.
The results are shown in Tab.~\ref{tab:baseline}.
Our method notably enhances the performance of all three models, with the most significant improvement seen in the segmentation task (38.81 \emph{vs.} 40.23).
The performance gain varies for different MIM methods, which depends on the choice of their original masking strategies and downstream tasks.
For models originally utilizing random masking, such as MAE~\cite{he2022masked} and SimMIM~\cite{xie2022simmim}, our approach effectively boosts performance by explicitly capturing improved relationships between different parts.

\textbf{Comparison with recent SSL methods.}
Comparison with recent SSL methods on common imageNet-1K classification settings are shown in Tab.~\ref{tab:sota}.
Related work ADIOS~\cite{shi2022adversarial}  is not included as it evaluates on other benchmarks.
DINO~\cite{caron2021emerging}, MoCo v3~\cite{chen2021empirical} and AttnMask~\cite{kakogeorgiou2022hide} use an extra momentum encoder as the teacher.
BEiT~\cite{bao2021beit} uses an extra $250M$ DALL-E data to pre-train dVAE tokenizer~\cite{ramesh2021zero}.
MST~\cite{li2021mst} introduces an MLP head to align the features of the teacher and student.
SemMAE~\cite{li2022semmae} uses a pre-trained iBOT to extract token features.
Our method gets comparable performance with fewer pre-training epochs, e.g., 200 \emph{v.s.} 300 in fine-tuning accuracy with state-of-art works.
With significantly less computation burden, the proposed method outperforms the SOTA by 1.1\%.
While Our method performs modestly on linear probing, the result shows linear probing and fine-tuning results are somewhat uncorrelated.
The evolved masking guides the network to learn more linearly separable features.
Since MIM methods are generally stronger non-linear feature extractors but perform unsatisfied on linear probing, we focus this work on improving fine-tune performance, i.e., a series of good initialization parameters for downstream tasks, as demonstrated in \cite{chen2021exploring,he2022masked}.

\textbf{Comparison with recent visual tokenizers.}
Recently, based on the remarkable inferring ability of LLMs, there has been a surge of research to combine LLMs and visual encoders for vision-language and even pure vision tasks.
This approach, using visual encoders as the visual tokenizer for LLMs, has also become an indicator to evaluate how well visual models can extract visual information.
In Tab.~\ref{tab:tokenizer}, we use GVTBench to compare our method's performance with popular visual tokenizers on various downstream tasks using the same architecture.
The compared tokenizers involve supervised models (DeiT~\cite{touvron2021training}), text-guided weakly supervised models (~\cite{radford2021learning}), and self-supervised models (MAE~\cite{he2022masked}, DINO~\cite{caron2021emerging}, DINOv2~\cite{oquab2023dinov2}).
The evaluated tasks include Visual Question Answering (VQA), Object Counting (OC), and Multi-Class Identification (MCI).
These tasks respectively gauge the model's visual comprehension from two important perspectives: grasping semantic understanding and perceiving fine-grained visual details. VQA emphasizes the tokenizer's ability to grasp high-level semantics, while OC and MCI assess its perception of intricate visual details.
During training, the language model remains frozen, while the visual tokenizer can either be frozen or optimized jointly.

The results in Tab.~\ref{tab:tokenizer} show that the proposed method narrows the gap between ImageNet-1K self-supervised techniques and large-scale pre-training. For the VQA task, the text-guided weakly supervised pre-trained model excels due to its use of extensive data for enhanced semantic understanding. CLIP surpasses supervised Deit, self-supervised DINO, and MAE by 2.73\%, 2.14\%, and 3.33\% without joint tuning. The proposed method built upon MAE, minimizing the gap to 1.79\%. With joint tuning of visual tokenizers, this gap is further reduced to 0.60\%.
This is because the proposed evolved hierarchy masking method targets training at various visual modeling levels.
In this way, even on small-scale pre-training data, we can extract a large number of visual dependencies, making pre-training more efficient.
On the other hand, the method enhances the self-supervised approach's capability to perceive intricate details.
The inherent nature of supervised and self-supervised training determines their finesse in extracting detailed information over text-guided weakly supervised.
The proposed method bolsters this advantage, boosting MAE's OC COCO, MCI COCO, and MCI VCR performance by 1.30\%, 0.39\%, and 0.83\% respectively.

\begin{table}[]
  \centering
  \small
  \caption{Segmentation performance on ADE20K~\cite{zhou2017scene} using the UperNet~\cite{xiao2018unified} framework.
  The backbone is initialized with DeiT~\cite{touvron2021training}, MAE~\cite{he2022masked}, CLIP~\cite{radford2021learning} and our pre-trained model on ViT-Base model, respectively.
  }\label{tab:seg}
  \begin{tabular}{cccccc}
    \toprule
  Method & Supervised &  \begin{tabular}[c]{@{}c@{}}\# Pretraining\\ Images\end{tabular}  &mIoU        &  aAcc  & mAcc  \\ \hline
  DeiT~\cite{touvron2021training}   & Fully      &  1.28M   &40.11 &  80.22  & 51.38 \\
  MAE~\cite{he2022masked}    & Self       &  1.28M &41.24 &  80.62  & 52.77 \\
  CLIP~\cite{radford2021learning}   & Weakly     &  400M  &35.54 &   77.65 & 46.56 \\
  Ours   & Self       &  1.28M  &47.26 &  81.97  & 58.92 \\  \bottomrule
  \end{tabular}
  \end{table}

\begin{figure*}[t]
    \centering
    \includegraphics[width=0.95\linewidth]{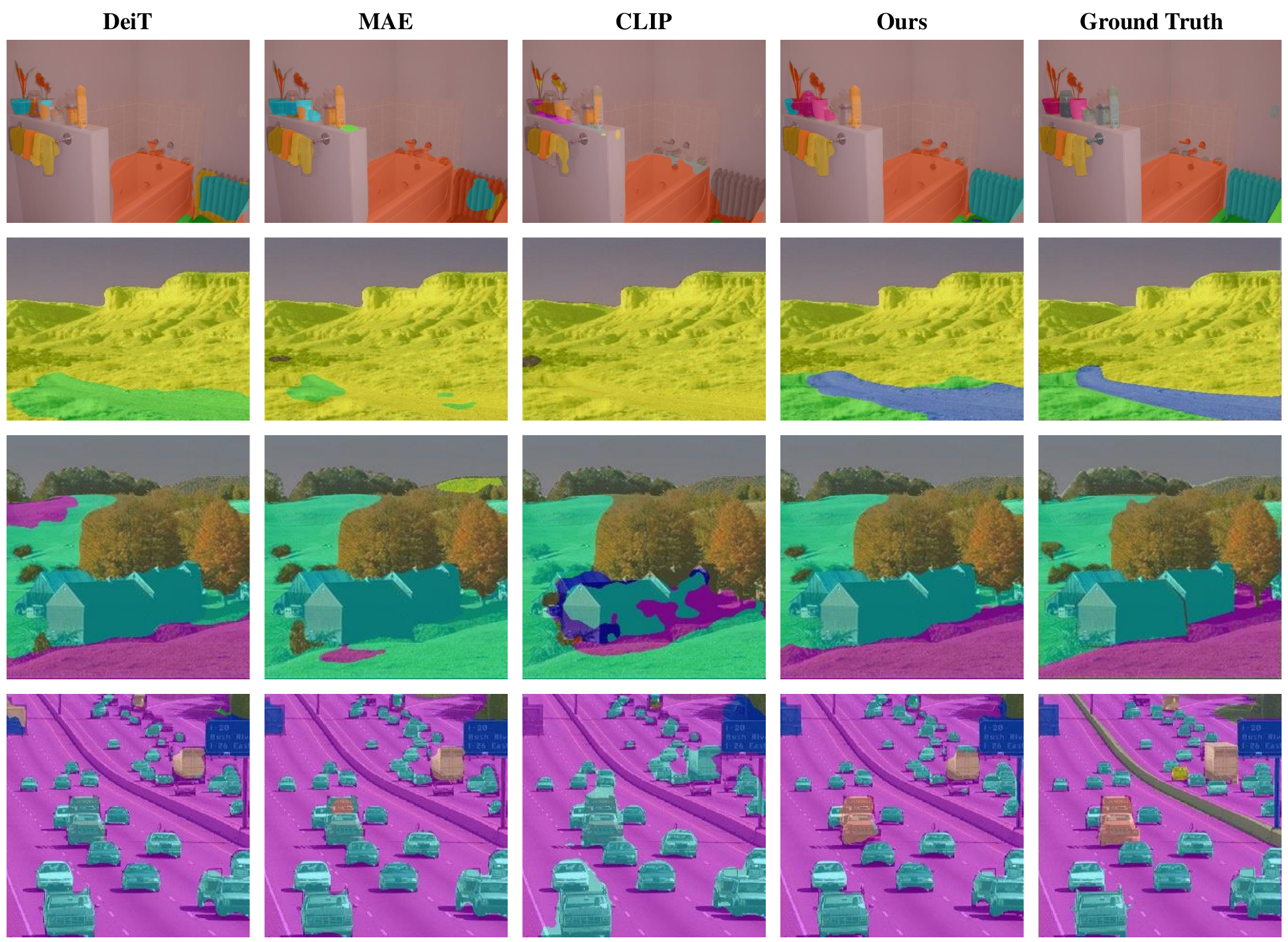}
    \caption{\textbf{Visualized segmentation results} on ADE20K~\cite{zhou2017scene} with DeiT~\cite{touvron2021training}, MAE~\cite{he2022masked}, CLIP~\cite{radford2021learning}, and the proposed method pretrained, respectively. Different colors denote different objects. Best viewed in color pdf.}
    \label{fig:seg}
  \end{figure*}

\textbf{Comparison on dense tasks.}
Tab.~\ref{tab:seg} reports the segmentation performance on ADE20K~\cite{zhou2017scene}.
The compared  methods includes popular fully supervised DeiT~\cite{touvron2021training}, self-supervised MAE~\cite{he2022masked} and large-scale weakly supervised CLIP~\cite{radford2021learning}.
We take these pre-trained models to replace the backbone in the UperNet~\cite{xiao2018unified} framework.
Compared to the commonly used fully supervised backbone and the baseline method MAE, our approach outperforms them by  7.15\% and 6.02\% mIoU respectively.
CLIP, despite being trained weakly at a large scale, does not perform well after fine-tuning.
We further visualize segmentation results of UperNet~\cite{xiao2018unified} in Fig.~\ref{fig:seg}.
Our method excels at accurately identifying segmentation categories.
For example, in the second column of images, our method effectively recognizes the road, while other methods incorrectly identify it as land.
It can be seen the proposed hierarchical masking method effectively enhances the pre-trained model to recognize texture and semantics, making it advantageous for dense prediction tasks.

\begin{table}[t!]
  \centering
  \small
  \caption{Landmark retrieval performance on Oxford and Paris dataset. We report mean average precision (mAP) with both Medium and Hard evaluation protocols.
  }\label{tab:ret}
  \begin{tabular}{cccccc}
    \toprule
  \multirow{2}{*}{Model}& \multirow{2}{*}{Supervised} & \multicolumn{2}{c}{Medium} & \multicolumn{2}{c}{Hard} \\ \cmidrule(r){3-4} \cmidrule(r){5-6}
             &            & $\mathcal{R}$Oxf         & $\mathcal{R}$Par        & $\mathcal{R}$Oxf        & $\mathcal{R}$Par       \\ \hline
  DeiT~\cite{touvron2021training}   &     Fully      & 56.17        & 72.40        & 31.09       & 54.46      \\
  CLIP~\cite{radford2021learning}    &     Weakly      & 45.42         & 58.68       & 15.32       & 39.22      \\
  Dino~\cite{caron2021emerging}      &     Self   & 51.53        & 75.13       & 24.15       & 52.03      \\
  MAE~\cite{he2022masked}            &    Self  & 55.38         & 76.55       & 30.74       & 50.06      \\
  Ours          &   Self   & 59.06        & 79.43       & 33.24       & 58.87     \\ \bottomrule
  \end{tabular}
  \end{table}

  \begin{figure}[t]
    \centering
    \small
     \includegraphics[width=1\linewidth]{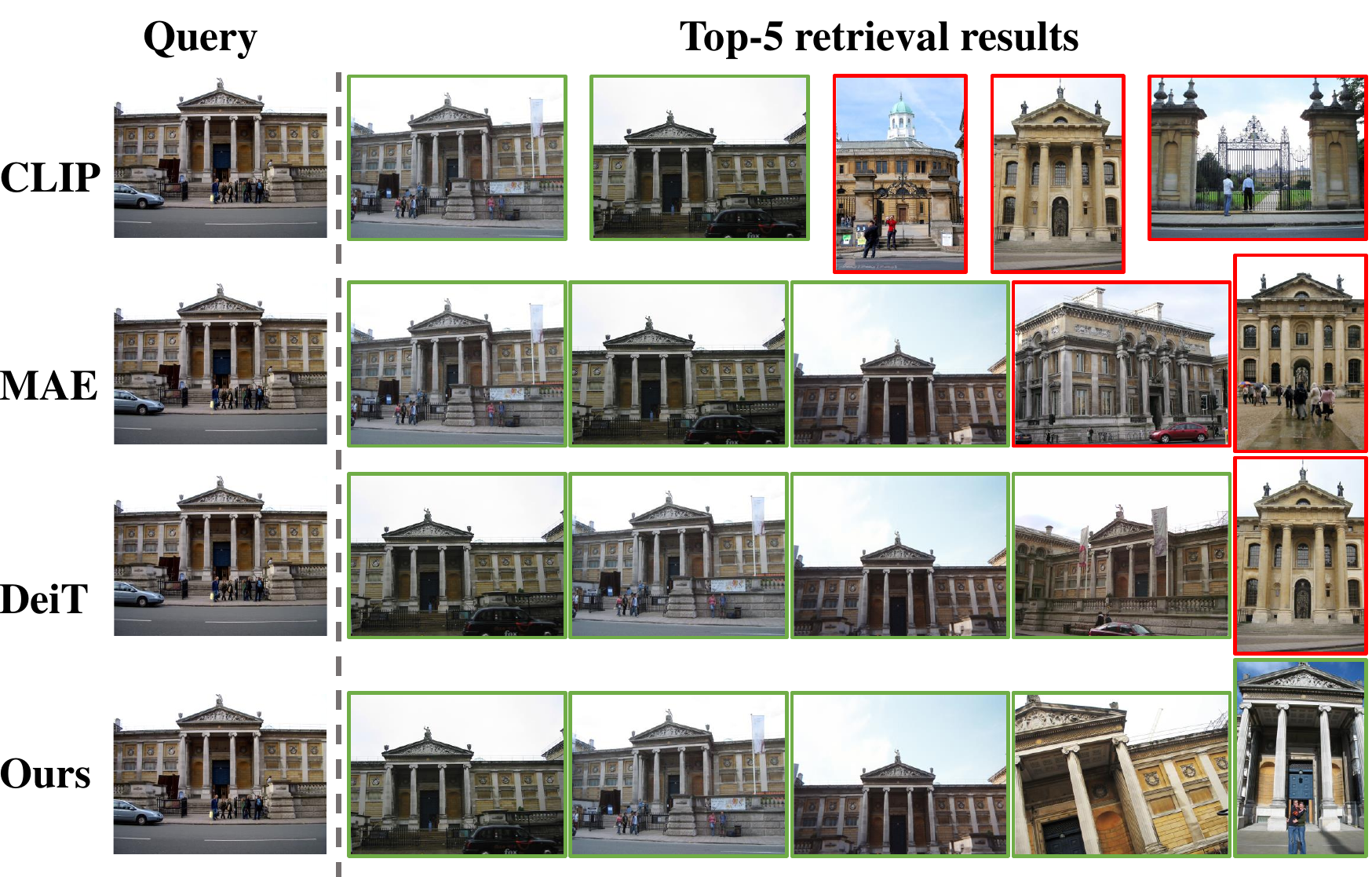}

     \caption{\textbf{Visualization of top-5 retrieval results} on Oxford dataset~\cite{philbin2007object}. Green and red boxes indicate correct and wrong predictions, respectively. }
     \label{fig:retrieval}
  \end{figure}

\textbf{Comparison on landmark retrieval.}
Tab.~\ref{tab:ret} compares our model against the popular pretraining methods on a fine-grained visual understanding task, landmark retrieval.
We present Mean Average Precision (mAP) results for both Medium and Hard splits of the Oxford and Paris image retrieval datasets.
Our models are finetuned on GLDv2 for 6M steps following DELG. Dino is pre-trained on a 1.2M clean set of GLDv2.
It can be seen that self-supervised and supervised methods outperform large-scale weakly supervised CLIP on fine-grained perception.
The proposed approach significantly enhances the fine-grained perception ability of pre-trained models, demonstrating the effectiveness of hierarchical masking in learning texture details.

\begin{figure}[t]
  \centering
   \includegraphics[width=0.95\linewidth]{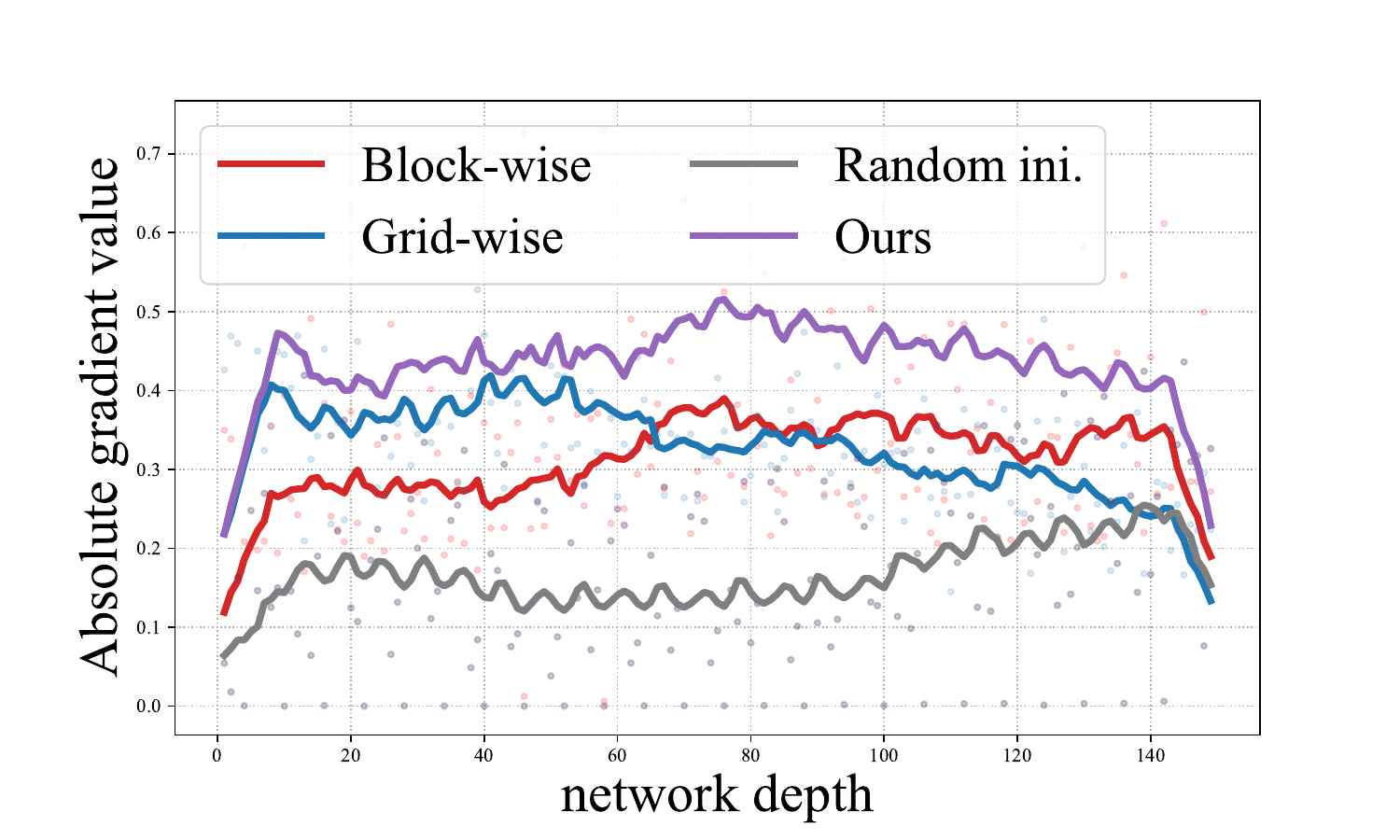}
   \caption{\textbf{Fine-tuning gradient value against network depth} on imageNet-1K classification~\cite{deng2009imagenet}. The data are fitted using exponential moving averages for better visualization. The gray line represents the model with random initialization of parameters, without pre-training. }
   \label{fig:grad}
\end{figure}

\subsection{Discussion}\label{sec:analysis}
This section studies the properties of masking methods and analyses the causes of their performance gap.

\textbf{Training costs.}
We measured both the space and time consumption of the proposed masking strategy during pretraining. The results show that mask generation occupies only 1.3\% of the total memory space and 8.1\% of the pretraining time. Specifically, obtaining the attention map accounts for 7.9\% of the total computation time, while constructing the binary tree takes up only 0.2\%. In our implementation, attention map $A$ is acquired by averaging the outputs of the initial layers, i.e., the first 6 layers in our experiments, instead of fully engaging the entire model depth. Furthermore, we share the features after the embedding layer to avoid redundant computations. The experimental results demonstrate the efficiency of the proposed method. Compared to the saved training epochs, the increased time consumption is minor.

\textbf{Fine-tuning gradient.}
The lottery ticket hypothesis~\cite{frankle2018lottery} demonstrates winning ticket weights tend to change by a larger amount than weights in the rest of the network, which is accompanied by larger gradients. Fig.~\ref{fig:grad} shows the fine-tuning gradients for different initialized models.  It can be seen that grid-wise masks better help model convergence in shallow layers.
And block-wise masking is more helpful for training deep layers, promoting the model learning high-level semantic relationships.
The proposed evolved masking method facilitates both deep and shallow network layers.

To summarize, the visual hints shared in visible and masked patches determine the level of abstraction at which visual knowledge is learned.
The masks where visible and masked patches with similar hints make the pre-trained model learn more low-level texture knowledge and better facilitate model convergence in shallow layers during fine-tuning.
Masking visual cues at a higher abstraction level prompts the model to pay more attention to global information and learn the connection between objects, which benefits more on the convergence of deep layers and dense prediction tasks.
The proposed method allows masks to adapt with model training, facilitating effective learning across various visual modeling levels, and providing stronger initialization parameters for pretraining-agnostic downstream tasks.

\section{Conclusion}
This paper proposes to enhance existing MIM methods by designing an evolved hierarchy masking strategy.
We found that the design of masks directly determines the visual knowledge learned by the network in each iteration.
The proposed method leverages the current training network to analyze images, capturing relationships among potential visual cues without requiring an additional network. Subsequently, it selectively masks patches corresponding to some chosen cues.
Through this technique, we can extract abundant visual dependencies from a limited pre-training dataset, making pre-training more focused and effective.
Experimental results demonstrate that the proposed method effectively improves the efficiency of pre-training, and has achieved performance improvements in seven downstream tasks.
When combined with Large Language Models, the proposed method not only underscores the advantages of self-supervised pre-training but also narrows the gap compared to extensive pre-training for semantic tasks.
We hope that the proposed method can provide timely insights for a cost-effective pathway towards self-supervised large-scale pre-training.


\section*{Acknowledgements} 
This work is supported in part by Grant No. 2023-JCJQ-LA-001-088, in part by Natural Science Foundation of China under Grant No. U20B2052, 61936011, in part by the Okawa Foundation Research Award, in part by the Ant Group Research Fund, and in part by the Kunpeng\&Ascend Center of Excellence, Peking University.


\bibliographystyle{IEEEtran}
\bibliography{sample-base}

\vfill

\end{document}